\documentclass[10pt,twocolumn,letterpaper]{article}

\usepackage{cvpr}
\usepackage{times}
\usepackage{epsfig}
\usepackage{graphicx}
\usepackage{amsmath}
\usepackage{amssymb}

\usepackage{caption}
\usepackage{color}
\usepackage{array}
\usepackage{paralist}   
\usepackage{multirow}

\usepackage[pagebackref=true,breaklinks=true,letterpaper=true,colorlinks,bookmarks=false]{hyperref}

\definecolor{grey50}{rgb}{0.5,0.5,0.5}

\newcolumntype{C}[1]{>{\centering\let\newline\\\arraybackslash\hspace{0pt}}m{#1}}
\newcolumntype{L}[1]{>{\raggedright\let\newline\\\arraybackslash\hspace{0pt}}m{#1}}

\newcommand\Tstrut{\rule{0pt}{2.2ex}}       
\newcommand\Bstrut{\rule[-1.0ex]{0pt}{0pt}} 
\newcommand{\TBstrut}{\Tstrut\Bstrut}       

\cvprfinalcopy 

\def\cvprPaperID{****} 

\begin{document}

\title{Rethinking the Faster R-CNN Architecture for Temporal Action Localization}

\author{\makebox[0cm]{Yu-Wei Chao$^1$\thanks{Work done in part during an internship at Google Research.}\kern5pt, Sudheendra Vijayanarasimhan$^2$, Bryan Seybold$^2$, David A. Ross$^2$, Jia Deng$^1$, Rahul Sukthankar$^2$} \vspace{3mm} \\
\begin{minipage}{0.47\textwidth}
 \centering
 $^1$University of Michigan, Ann Arbor
 {\tt\small \{ywchao,jiadeng\}@umich.edu}\\
\end{minipage}
\begin{minipage}{0.53\textwidth}
 \centering
 $^2$Google Research
 {\tt\small \{svnaras,seybold,dross,sukthankar\}@google.com}\\
\end{minipage}
}

\maketitle

\begin{abstract}
We propose TAL-Net, an improved approach to temporal action localization in
video that is inspired by the Faster R-CNN object detection framework. TAL-Net
addresses three key shortcomings of existing approaches: (1) we improve
receptive field alignment using a multi-scale architecture that can accommodate
extreme variation in action durations; (2) we better exploit the temporal
context of actions for both proposal generation and action classification by
appropriately extending receptive fields; and (3) we explicitly consider
multi-stream feature fusion and demonstrate that fusing motion late is
important. We achieve state-of-the-art performance for both action proposal and
localization on THUMOS'14 detection benchmark and competitive performance on
ActivityNet challenge.
\end{abstract}

\section{Introduction}
\label{sec:intro}

Visual understanding of human actions is a core capability in building
assistive AI systems. The problem is conventionally studied in the setup of
action classification~\cite{wang:iccv2013,simonyan:nips2014,ng:cvpr2015}, where
the goal is to perform forced-choice classification of a temporally trimmed
video clip into one of several action classes. Despite fruitful progress, this
classification setup is unrealistic, because real-world videos are usually
untrimmed and the actions of interest are typically embedded in a background of
irrelevant activities. Recent research attention has gradually shifted to
temporal action localization in untrimmed
video~\cite{karaman:2014,oneata:2014,wang:2014}, where the task is to not only
identify the action class, but also detect the start and end time of each
action instance. Improvements in temporal action localization can drive
progress on a large number of important topics ranging from immediate
applications, such as extracting highlights in sports video, to higher-level
tasks, such as automatic video captioning.

Temporal action localization, like object detection, falls under the umbrella
of visual detection problems. While object detection aims to produce spatial
bounding boxes in a 2D image, temporal action localization aims to produce
temporal segments in a 1D sequence of frames. As a result, many approaches to
action localization have drawn inspiration from advances in object detection. A
successful example is the use of region-based
detectors~\cite{girshick:cvpr2014,girshick:iccv2015,ren:nips2015}. These
methods first generate a collection of class-agnostic region proposals from the
full image, and go through each proposal to classify its object class. To
detect actions, one can follow this paradigm by first generating segment
proposals from the full video, followed by classifying each proposal.

Among region-based detectors, Faster R-CNN~\cite{ren:nips2015} has been widely
adopted in object detection due to its competitive detection accuracy on public
benchmarks~\cite{lin:eccv2014,everingham:ijcv2015}. The core idea is to
leverage the immense capacity of deep neural networks (DNNs) to power the two
processes of proposal generation and object classification. Given its success
in object detection in images, there is considerable interest in employing
Faster R-CNN for temporal action localization in video. However, such a domain
shift introduces several challenges.
We review the issues of Faster R-CNN in the action localization domain, and
redesign the architecture to specifically address them. We focus on the
following:
\begin{compactenum}
 \vspace{0.45mm}
 \item \textbf{How to handle large variations in action durations?} The
temporal extent of actions varies dramatically compared to the size of objects
in an image---from a fraction of a second to minutes. However, Faster R-CNN
evaluates different scales of candidate proposals (i.e., anchors) based on a
shared feature representation, which may not capture relevant information due
to a misalignment between the temporal scope of the feature (i.e. receptive
field) and the span of the anchor. We propose to enforce such alignment using a
multi-tower network and dilated temporal convolutions.
 \vspace{0.45mm}
 \item \textbf{How to utilize temporal context?} The moments preceding and
following an action instance contain critical information for localization and
classification (arguably more so than the spatial context of an object). A
naive application of Faster R-CNN would fail to exploit this temporal context.
We propose to explicitly encode temporal context by extending the receptive
fields in proposal generation and action classification.
 \vspace{0.45mm}
 \item \textbf{How best to fuse multi-stream features?} State-of-the-art action
classification results are mostly achieved by fusing RGB and optical flow based
features. However, there has been limited work in exploring such feature fusion
for Faster R-CNN. We propose a late fusion scheme and empirically demonstrate
its edge over the common early fusion scheme.
\end{compactenum}
\vspace{0.45mm}
Our contributions are twofold: (1) we introduce the Temporal Action
Localization Network (TAL-Net), which is a new approach for action localization
in video based on Faster R-CNN; (2) we achieve state-of-the-art performance on
both action proposal and localization on the THUMOS'14 detection
benchmark~\cite{THUMOS14}, along with competitive performance on the
ActivityNet dataset~\cite{caba_heilbron:cvpr2015}.

\section{Related Work}

\paragraph{Action Recognition} Action recognition is conventionally formulated
as a classification problem. The input is a video that has been temporally
trimmed to contain a specific action of interest, and the goal is to classify
the action. Tremendous progress has recently been made due to the introduction
of large datasets and the developments on deep neural
networks~\cite{simonyan:nips2014,ng:cvpr2015,tran:iccv2015,wang:eccv2016,carreira:cvpr2017,feichtenhofer:cvpr2017}.
However, the assumption of trimmed input limits the application of these
approaches in real scenarios, where the videos are usually untrimmed and may
contain irrelevant backgrounds.

\vspace{-3mm}

\paragraph{Temporal Action Localization} Temporal action localization assumes
the input to be a long, untrimmed video, and aims to identify the start and end
times as well as the action label for each action instance in the video. The
problem has recently received significant research attention due to its
potential application in video data analysis. Below we review the relevant work
on this problem.

Early approaches address the task by applying temporal sliding windows followed
by SVM classifiers to classify the action within each
window~\cite{karaman:2014,oneata:2014,wang:2014,ni:cvpr2016,yuan:cvpr2016}.
They typically extract improved dense trajectory~\cite{wang:iccv2013} or
pre-trained DNN features, and globally pool these features within each window
to obtain the input for the SVM classifiers. Instead of global pooling, Yuan et
al.~\cite{yuan:cvpr2016} proposed a multi-scale pooling scheme to capture
features at multiple resolutions. However, these approaches might be
computationally inefficient, because one needs to apply each action classifier
exhaustively on windows of different sizes at different temporal locations
throughout the entire video.

Another line of work generates frame-wise or
snippet-wise action labels, and uses these labels to define the temporal
boundaries of
actions~\cite{ma:cvpr2016,singh:cvpr2016,dave:cvpr2017,lea:cvpr2017,yuan:cvpr2017,hou:bmvc2017}.
One major challenge here is to enable temporal contextual reasoning in
predicting the individual labels. Lea et al.~\cite{lea:cvpr2017} proposed novel
temporal convolutional architectures to capture long-range temporal
dependencies, while others~\cite{ma:cvpr2016,singh:cvpr2016,dave:cvpr2017} use
recurrent neural networks. A few other methods add a separate contextual
reasoning stage on top of the frame-wise or snippet-wise prediction scores to
explicitly model action durations or temporal
transitions~\cite{richard:cvpr2016,yuan:cvpr2017,hou:bmvc2017}.

Inspired by the recent success of region-based detectors in object
detection~\cite{girshick:cvpr2014,girshick:iccv2015}, many recent approaches
adopt a two-stage, proposal-plus-classification
framework~\cite{caba_heilbron:cvpr2016,shou:cvpr2016,escorcia:eccv2016,buch:cvpr2017,caba_heilbron:cvpr2017,shou:cvpr2017,zhao:iccv2017},
i.e. first generating a sparse set of class-agnostic segment proposals from the
input video, followed by classifying the action categories for each proposal.
A large number of these works focus on improving the segment
proposals~\cite{caba_heilbron:cvpr2016,escorcia:eccv2016,caba_heilbron:cvpr2017,buch:cvpr2017},
while others focus on building more accurate action
classifiers~\cite{shou:cvpr2017,zhao:iccv2017}. However, most of these methods
do not afford end-to-end training on either the proposal or classification
stage. Besides, the proposals are typically selected from sliding windows of
predefined scales~\cite{shou:cvpr2016}, where the boundaries are fixed and may
result in imprecise localization if the windows are not dense.


\begin{figure*}[t]
 \centering
 \begin{minipage}{0.48\textwidth}
  \centering
  \includegraphics[width=0.85\linewidth]{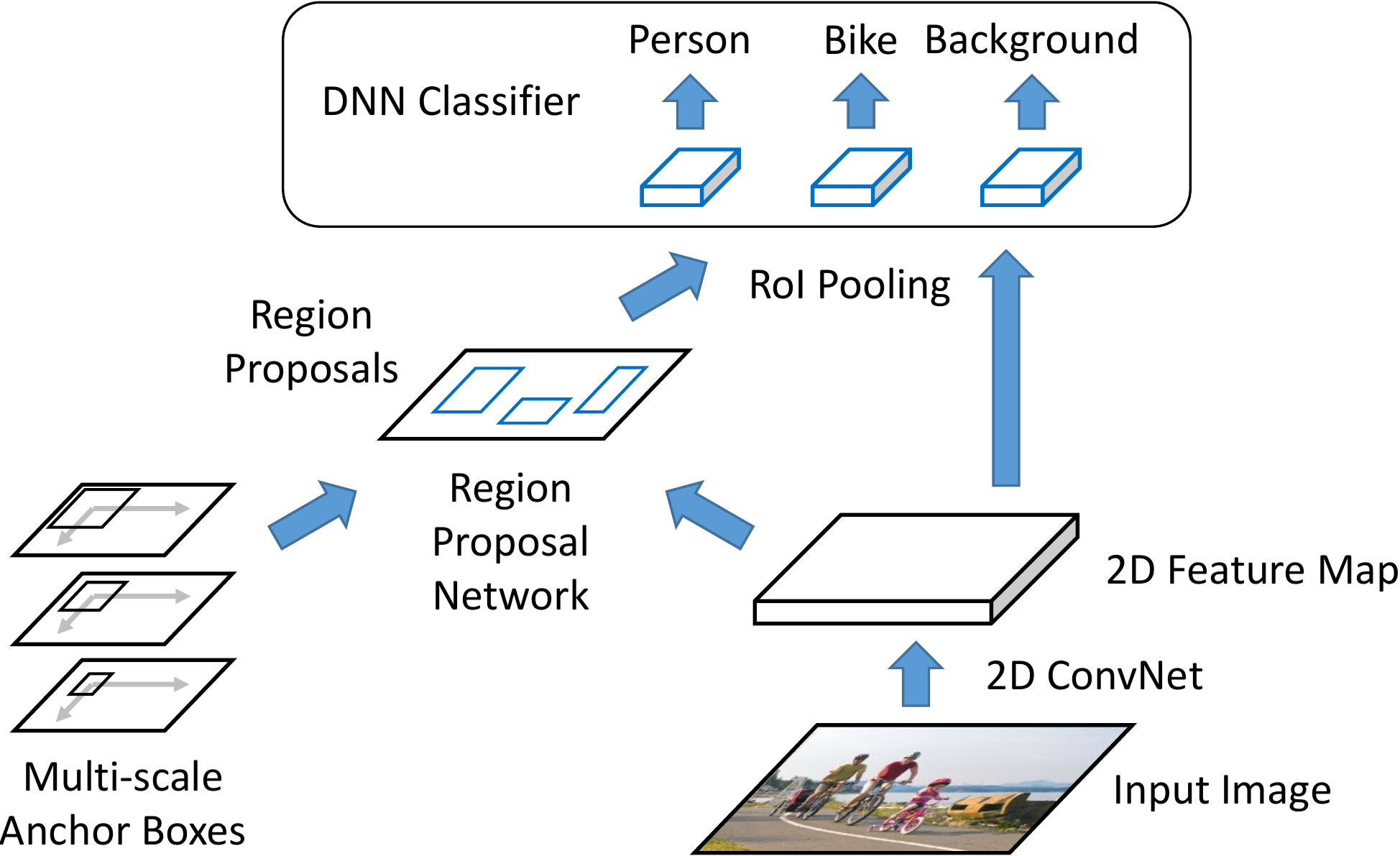}
 \end{minipage}
 \begin{minipage}{0.48\textwidth}
  \centering
  \includegraphics[width=1.00\linewidth]{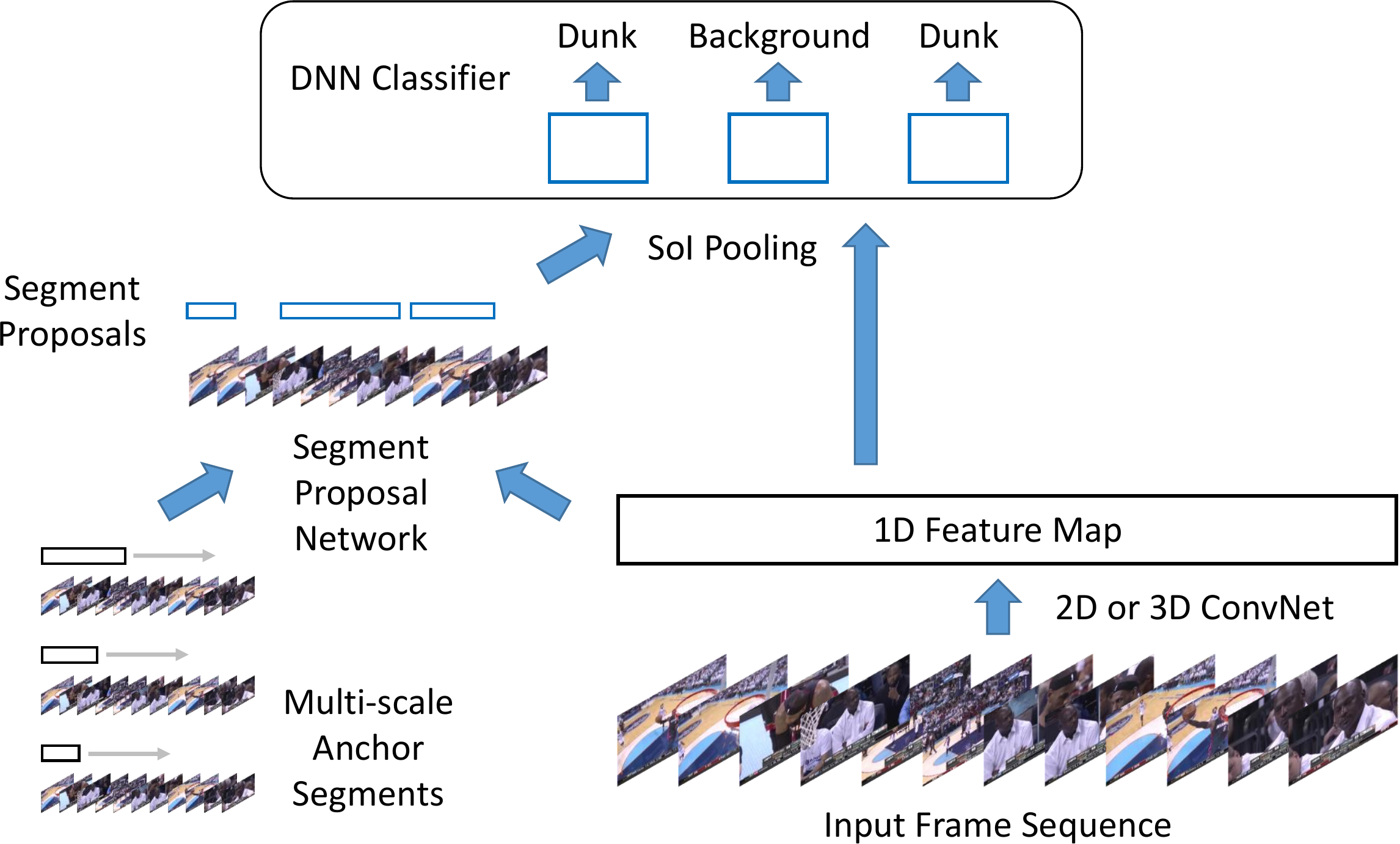}
 \end{minipage}
 \caption{\small Contrasting the Faster R-CNN architecture for object detection
in images~\cite{ren:nips2015} (left) and temporal action localization in
video~\cite{gao:bmvc2017,dai:iccv2017,gao:iccv2017,xu:iccv2017} (right).
Temporal action localization can be viewed as the 1D counterpart of the object
detection problem.}
 \label{fig:faster_r-cnn}
\end{figure*}

As the latest incarnation of the region-based object detectors, the Faster
R-CNN architecture~\cite{ren:nips2015} is composed of end-to-end trainable
proposal and classification networks, and applies region boundary regression in
both stages. A few very recent works have started to apply such architecture to
temporal action
localization~\cite{gao:bmvc2017,dai:iccv2017,gao:iccv2017,xu:iccv2017}, and
demonstrated competitive detection accuracy. In particular, the R-C3D
network~\cite{xu:iccv2017} is a classic example that closely follows the
original Faster R-CNN in many design details. While being a powerful detection
paradigm, we argue that naively applying the Faster R-CNN architecture to
temporal action localization might suffer from a few issues. We propose to
address these issues in this paper. We will also clarify our contributions over
other Faster R-CNN based
methods~\cite{gao:bmvc2017,dai:iccv2017,gao:iccv2017,xu:iccv2017} later when we
introduce TAL-Net.

In addition to the works reviewed above, there exist other classes of
approaches, such as those based on single-shot
detectors~\cite{buch:bmvc2017,lin:acmmm2017} or reinforcement
learning~\cite{yeung:cvpr2016}. Others have also studied temporal action
localization in a weakly supervised setting~\cite{sun:acmmm2015,wang:cvpr2017},
where only video-level action labels are available for training. Also note that
besides temporal action localization, there also exists a large body of work on
spatio-temporal action
localization~\cite{gkioxari:cvpr2015,kalogeiton:iccv2017,singh:iccv2017}, which
is beyond the scope of this paper.

\section{Faster R-CNN}

We briefly review the Faster R-CNN detection framework in this section. Faster
R-CNN is first proposed to address object detection~\cite{ren:nips2015}, where
given an input image, the goal is to output a set of detection bounding boxes,
each tagged with an object class label. The full pipeline consists of two
stages: \textit{proposal generation} and \textit{classification}. First, the
input image is processed by a 2D ConvNet to generate a 2D feature map. Another
2D ConvNet (referred to as the Region Proposal Network) is then used to
generate a sparse set of class-agnostic region proposals, by classifying a
group of scale varying anchor boxes centered at each pixel location of the
feature map. The boundaries of the proposals are also adjusted with respect to
the anchor boxes through regression. Second, for each region proposal, features
within the region are first pooled into a fixed size feature map (i.e. RoI
pooling~\cite{girshick:iccv2015}). Using the pooled feature, a DNN classifier
then computes the object class probabilities and simultaneously regresses the
detection boundaries for each object class. Fig.~\ref{fig:faster_r-cnn} (left)
illustrates the full pipeline. The framework is conventionally trained by
alternating between the training of the first and second
stage~\cite{ren:nips2015}.

Faster R-CNN naturally extends to temporal action
localization~\cite{gao:bmvc2017,dai:iccv2017,xu:iccv2017}. Recall that object
detection aims to detect 2D spatial regions, whereas in temporal action
localization, the goal is to detect 1D temporal \textit{segments}, each
represented by a \textit{start} and an \textit{end} time. Temporal action
localization can thus be viewed as the 1D counterpart of object detection. A
typical Faster R-CNN pipeline for temporal action localization is illustrated
in Fig.~\ref{fig:faster_r-cnn} (right). Similar to object detection, it
consists of two stages. First, given a sequence of frames, we extract a 1D
feature map, typically via a 2D or 3D ConvNet. The feature map is then passed
to a 1D ConvNet~\footnote{``1D convolution'' \& ``temporal convolution'' are
used interchangeably.} (referred to as the Segment Proposal Network) to
classify a group of scale varying anchor segments at each temporal location,
and also regress their boundaries. This returns a sparse set of class-agnostic
segment proposals. Second, for each segment proposal, we compute the action
class probabilities and further regress the segment boundaries,
by first applying a 1D RoI pooling (termed ``SoI pooling'') layer followed by a
DNN classifier.

\section{TAL-Net}

TAL-Net follows the Faster R-CNN detection paradigm for temporal action
localization (Fig.~\ref{fig:faster_r-cnn} right) but features three novel
architectural changes (Sec.~\ref{sec:receptive} to~\ref{sec:fusion}).


\subsection{Receptive Field Alignment}
\label{sec:receptive}

\begin{figure*}[t]
 \centering
 \begin{minipage}{0.48\textwidth}
  \centering
  \includegraphics[width=0.98\linewidth]{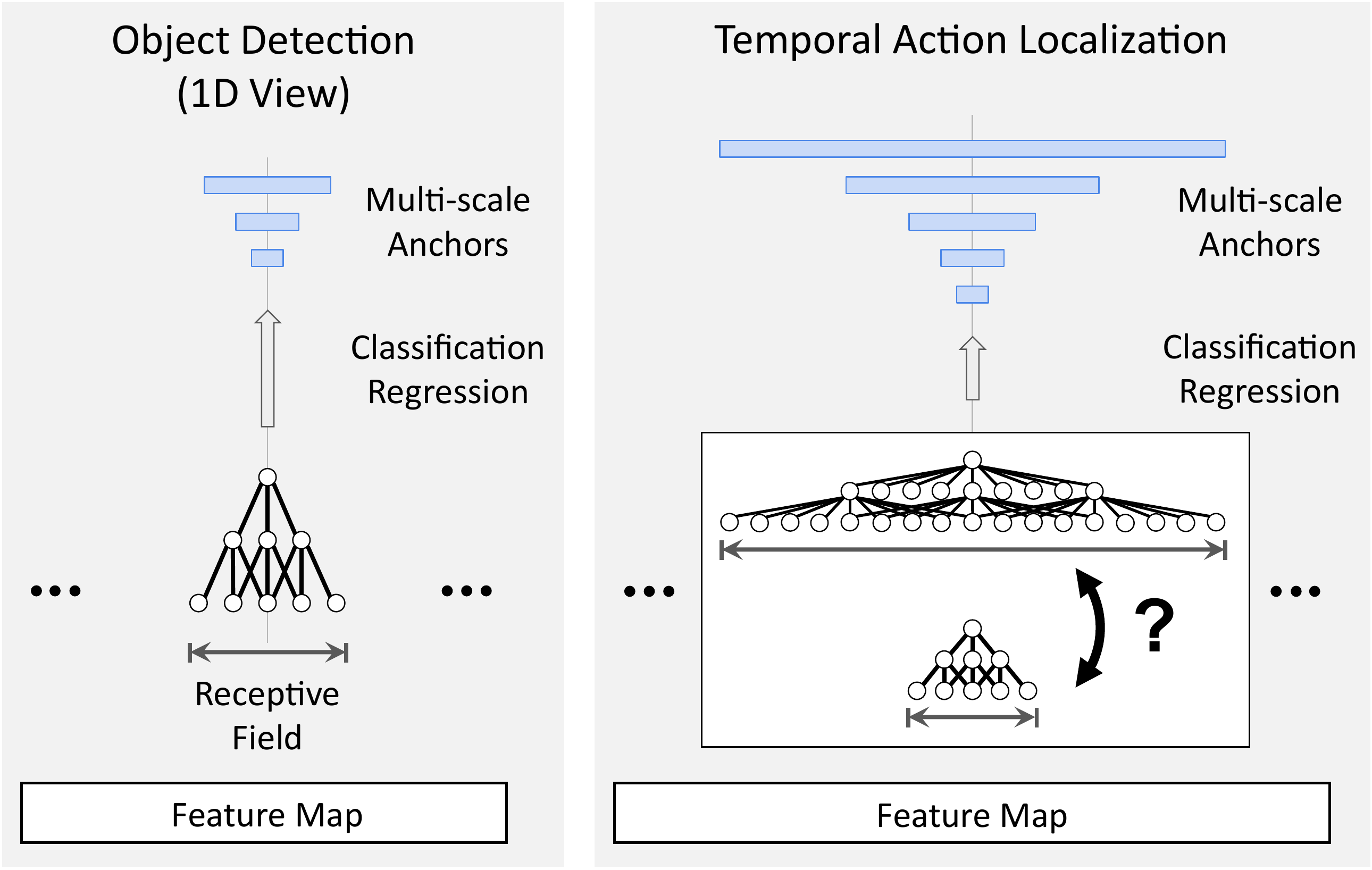}
 \end{minipage}
 \hspace{5mm}
 \begin{minipage}{0.48\textwidth}
  \centering
  \includegraphics[width=0.98\linewidth]{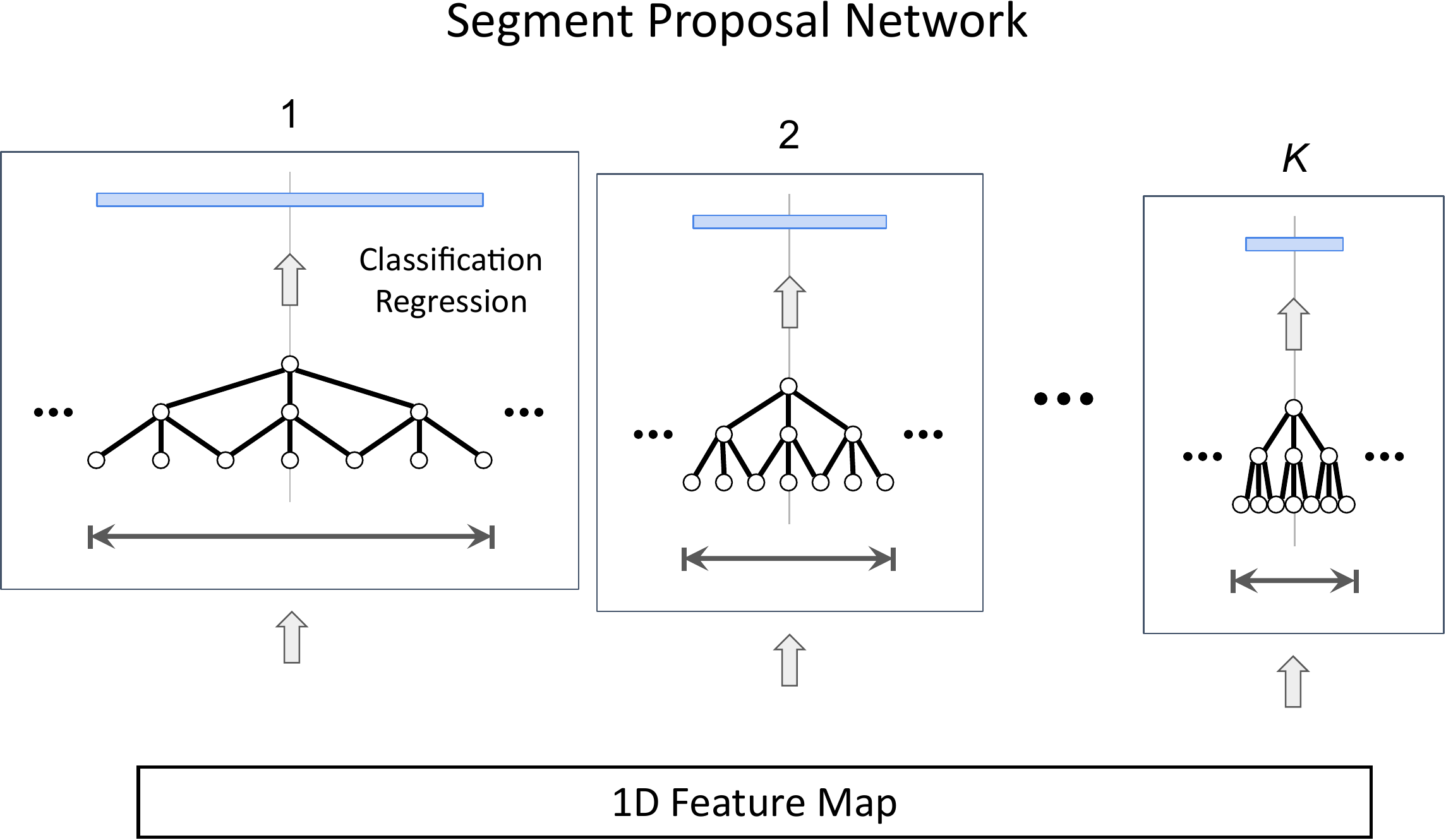}
 \end{minipage}
 \vspace{-1mm}
 \caption{\small Left: The limitation of sharing the receptive field across
different anchor scales in temporal action localization. Right: The multi-tower
architecture of our Segment Proposal Network. Each anchor scale has an
associated network with aligned receptive field.}
 \vspace{-2mm}
 \label{fig:multi-tower}
\end{figure*}

\begin{figure}[t]
 \centering
 \includegraphics[width=0.96\linewidth]{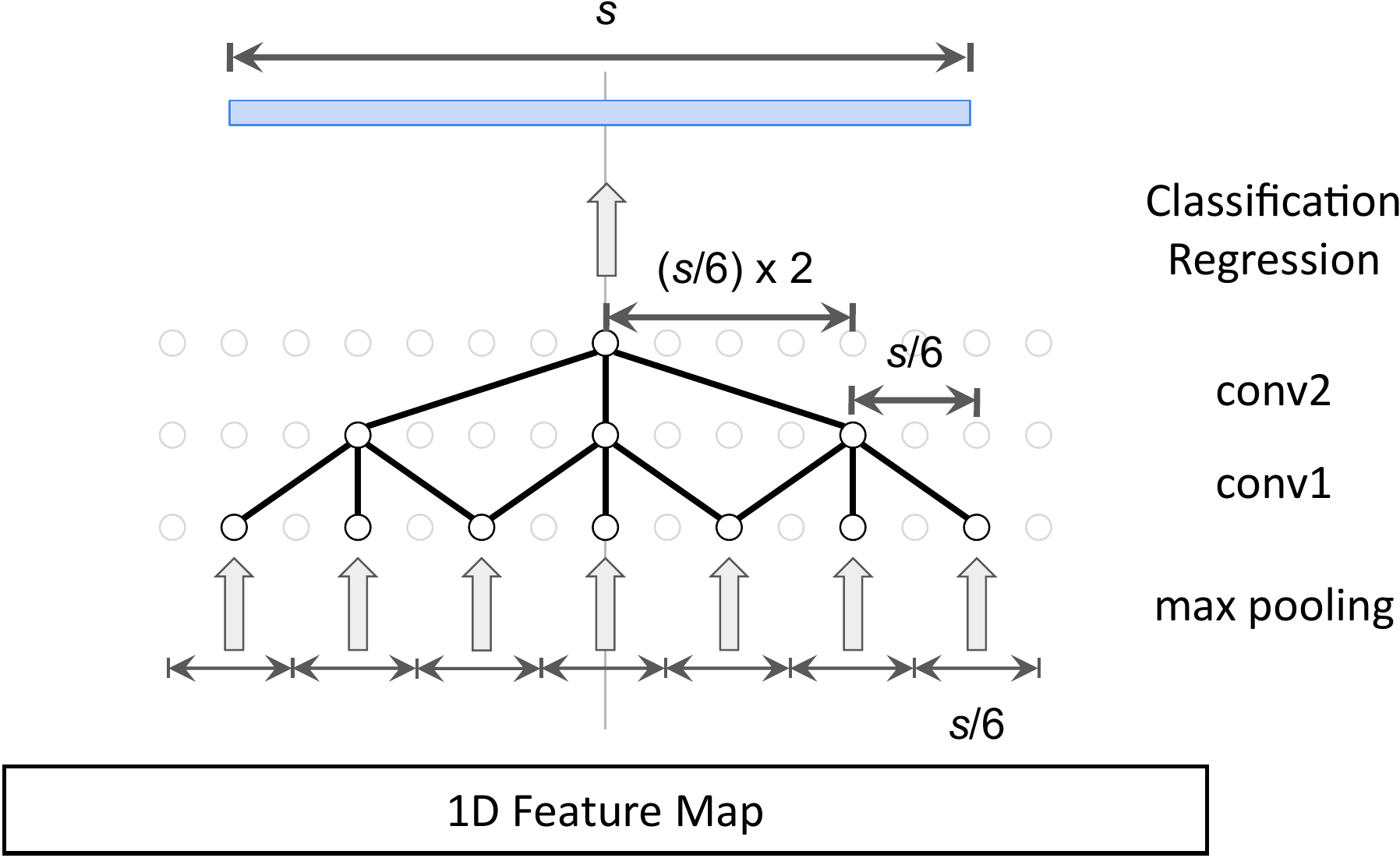}
 \vspace{-1mm}
 \caption{\small Controlling the receptive field size $s$ with dilated temporal
convolutions.}
 \vspace{-2mm}
 \label{fig:dilated_conv}
\end{figure}

Recall that in proposal generation, we generate a sparse set of class-agnostic
proposals by classifying a group of scale varying anchors at each location in
the feature map. In object detection~\cite{ren:nips2015}, this is achieved by
applying a small ConvNet on top of the feature map, followed by a $1\times1$
convolutional layers with $K$ filters, where $K$ is the number of scales. Each
filter will classify the anchor of a particular scale. This reveals an
important \textit{limitation}: the anchor classifiers at each location share
the same receptive field. Such design may be reasonable for object detection,
but may not generalize well to temporal action localization, because the
temporal length of actions can vary more drastically compared to the spatial
size of objects, e.g. in THUMOS'14~\cite{THUMOS14}, the action lengths range
from less than a second to more than a minute. To ensure a high recall, the
applied anchor segments thus need to have a wide range of scales
(Fig.~\ref{fig:multi-tower} left). However, if the receptive field is set too
small (i.e. temporally short), the extracted feature may not contain sufficient
information when classifying large (i.e. temporally long) anchors, while if it
is set too large, the extracted feature may be dominated by irrelevant
information when classifying small anchors.


To address this issue, we propose to align each anchor's receptive field with
its temporal span. This is achieved by two key enablers: a \textit{multi-tower}
network and \textit{dilated temporal convolutions}. Given a 1D feature map, our
Segment Proposal Network is composed of a collection of $K$ temporal ConvNets,
each responsible for classifying the anchor segments of a particular scale
(Fig.~\ref{fig:multi-tower} right). Most importantly, each temporal ConvNet is
carefully designed such that its receptive field size coincides with the
associated anchor scale. At the end of each ConvNet, we apply two parallel
convolutional layers with kernel size 1 for anchor classification and boundary
regression, respectively.

The next question is: how do we design temporal ConvNets with a controllable
receptive field size $s$? Suppose we use temporal convolutional filters with
kernel size 3 as a building block. One way to increase $s$ is simply stacking
the convolutional layers: $s=2L+1$ if we stack $L$ layers. However, given a
target receptive field size $s$, the required number of layers $L$ will then
grow linearly with $s$, which can easily increase the number of parameters and
make the network prone to overfitting. One solution is to apply pooling layers:
if we add a pooling layer with kernel size 2 after each convolutional layer,
the receptive field size is then given by $s=2^{(L+1)}-1$. While now $L$ grows
logarithmically with $s$, the added pooling layers will exponentially reduce
the resolution of the output feature map, which may sacrifice localization
precision in detection tasks.

To avoid overgrowing the model while maintaining the resolution, we propose to
use dilated temporal convolutions. Dilated
convolutions~\cite{chen:iclr2015,yu:iclr2016} act like regular convolutions,
except that one subsamples pixels in the input feature map instead of taking
adjacent ones when multiplied with a convolution kernel. This technique has
been successfully applied to 2D ConvNets~\cite{chen:iclr2015,yu:iclr2016} and
1D ConvNets~\cite{lea:cvpr2017} to expand the receptive field without loss of
resolution. In our Segment Proposal Network, each temporal ConvNet consists of
only two dilated convolutional layers (Fig.~\ref{fig:dilated_conv}). To attain
a target receptive field size $s$, we can explicitly compute the required
dilation rate (i.e. subsampling rate) $r_l$ for layer $l$ by $r_1=s/6$ and
$r_2=(s/6)\times2$. We also smooth the input before subsampling by adding a max
pooling layer with kernel size $s/6$ before the first convolutional layer.

\vspace{-3mm}

\paragraph{Contributions
beyond~\cite{dai:iccv2017,gao:bmvc2017,gao:iccv2017,xu:iccv2017}} Xu et
al.~\cite{xu:iccv2017} followed the original Faster R-CNN and thus their
anchors at each pixel location still shared the receptive field. Both Gao et
al.~\cite{gao:bmvc2017,gao:iccv2017} and Dai et al.~\cite{dai:iccv2017} aligned
each anchor's receptive field with its span. However, Gao et
al.~\cite{gao:bmvc2017,gao:iccv2017} average pooled the features within the
span of each anchor, whereas we use temporal convolutions to extract
structure-sensitive features. Our approach is similar in spirit to Dai et
al.~\cite{dai:iccv2017}, which sampled a fixed number of features within the
span of each anchor; we approach this using dilated convolutions.


\subsection{Context Feature Extraction}
\label{sec:context}

Temporal context information (i.e. what happens immediately before and after an
action instance) is a critical signal for temporal action localization for two
reasons. First, it enables more accurate localization of the action boundaries.
For example, seeing a person standing still on the far end of a diving board is
a strong signal that he will soon start a ``diving'' action. Second, it
provides strong semantic cues for identifying the action class within the
boundaries. For example, seeing a javelin flying in the air indicates that a
person just finished a ``javelin throw'', not ``pole vault''. As a result, it
is critical to encode the temporal context features in the action localization
pipeline. Below we detail our approach to explicitly exploit context features
in both the proposal generation and action classification stage.

\begin{figure}[t]
 \centering
 \includegraphics[width=0.96\linewidth]{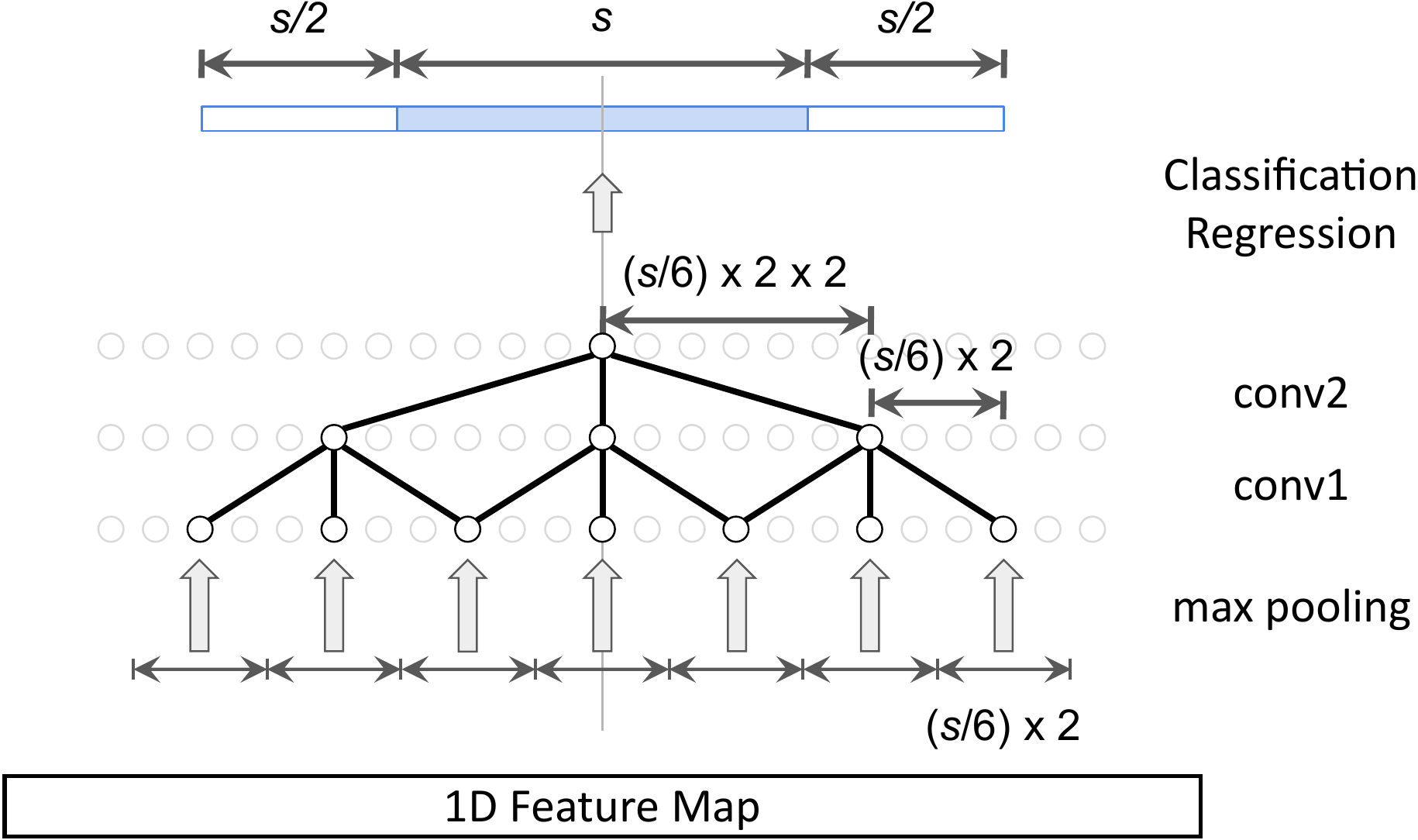}
 \vspace{-1mm}
 \caption{\small Incorporating context features in proposal generation.}
 \vspace{-2mm}
 \label{fig:context-proposal}
\end{figure}

\begin{figure}[t]
 \centering
 \begin{minipage}{0.48\textwidth}
  \centering
  \includegraphics[width=0.96\linewidth]{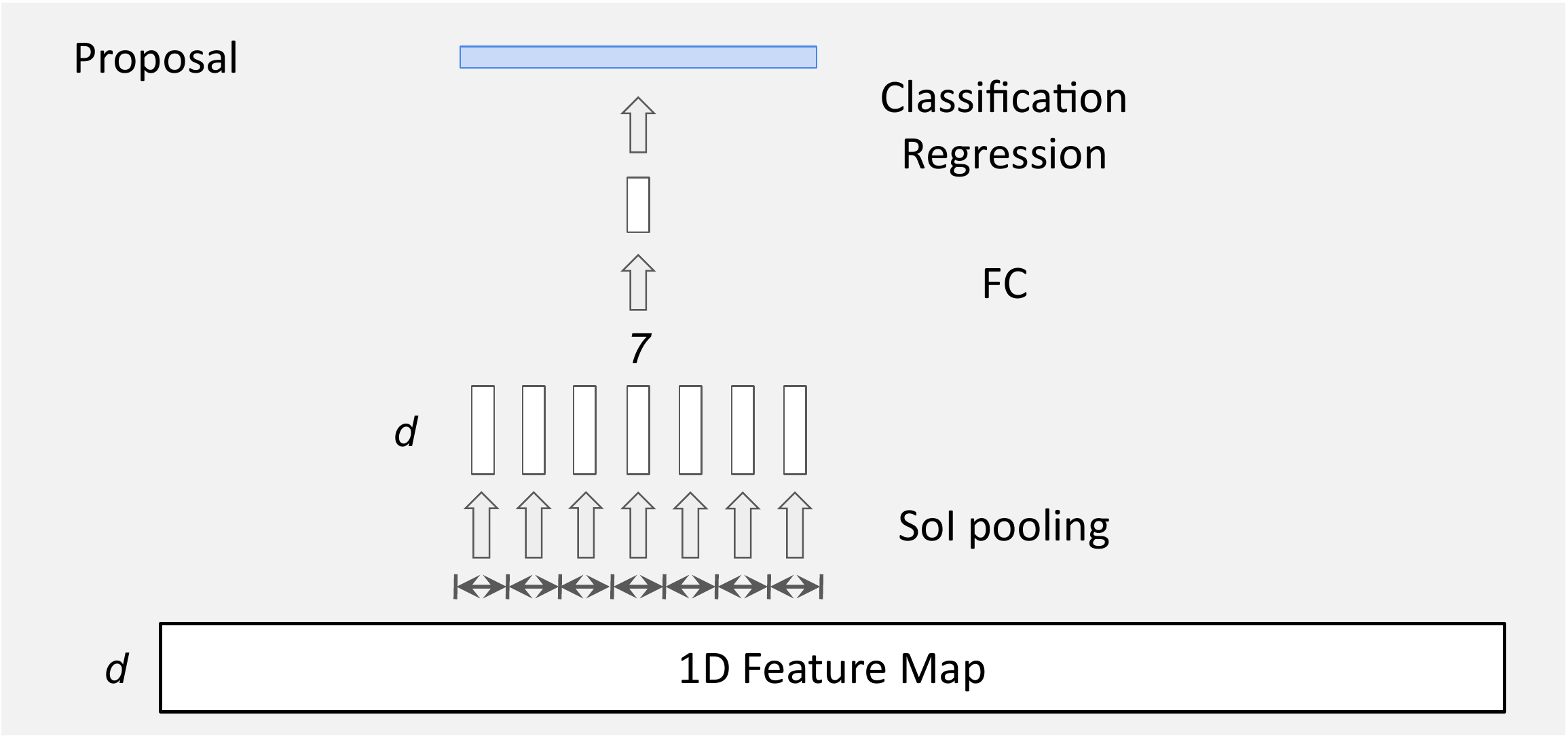}
 \end{minipage}
 \\\vspace{1mm}
 \begin{minipage}{0.48\textwidth}
  \centering
  \includegraphics[width=0.96\linewidth]{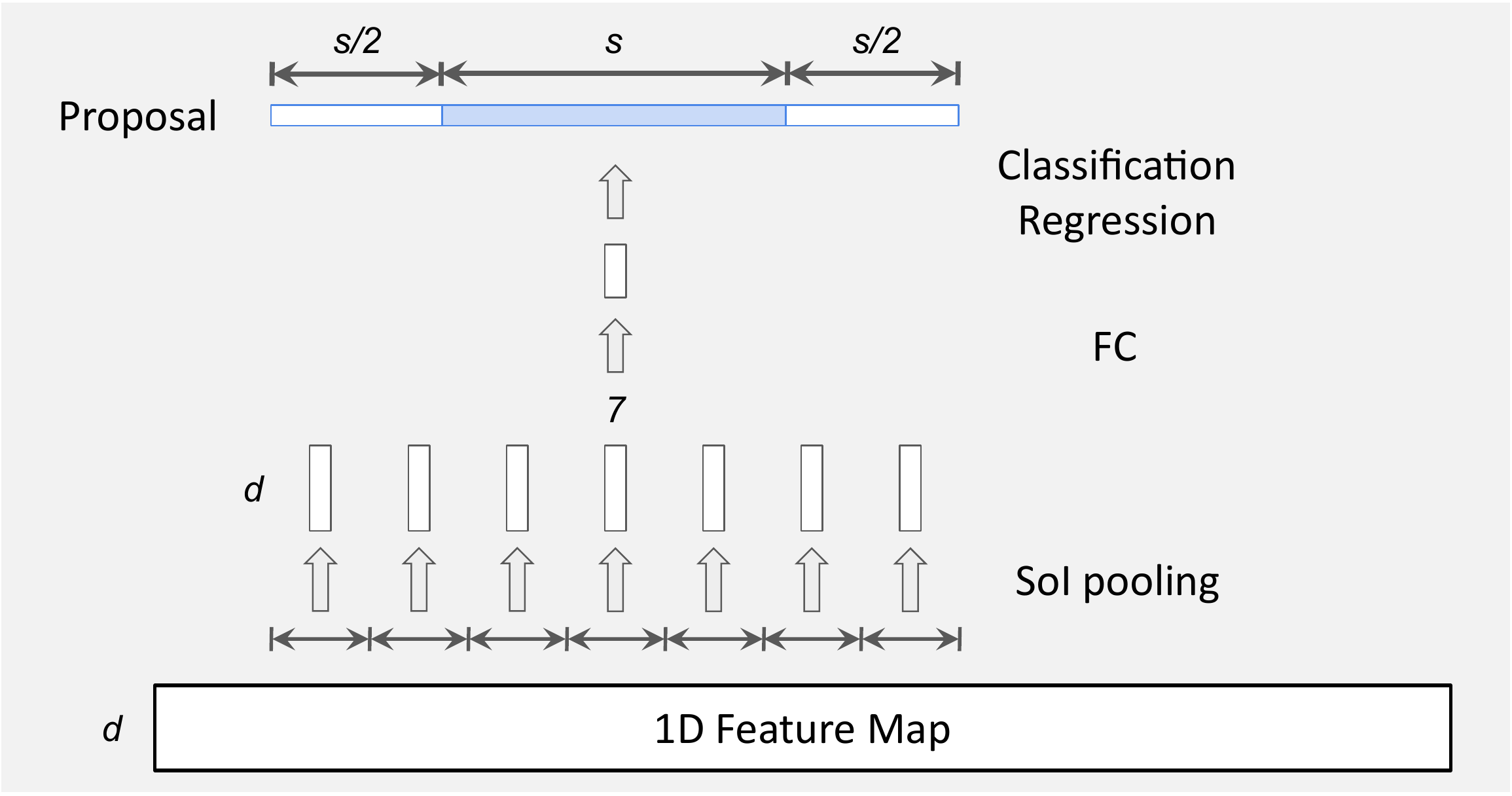}
 \end{minipage}
 \vspace{-1mm}
 \caption{\small Classifying a proposal without
(top)~\cite{girshick:iccv2015,ren:nips2015} and with (bottom) incorporating
context features}
 \vspace{-2mm}
 \label{fig:context-classify}
\end{figure}

In proposal generation, we showed the receptive field for classifying an anchor
can be matched with the anchor's span (Sec.~\ref{sec:receptive}). However, this
only extracts the features within the anchor, and overlooks the contexts before
and after it. To ensure the context features are used for anchor classification
and boundary regression, the receptive field must cover the context regions.
Suppose the anchor is of scale $s$, we enforce the receptive field to also
cover the two segments of length $s/2$ immediately before and after the anchor.
This can be achieved by doubling the dilation rate of the convolutional layers,
i.e. $r_1=(s/6)\times2$ and $r_2=(s/6)\times2\times2$, as illustrated in
Fig.~\ref{fig:context-proposal}. Consequently, we also double the kernel size
of the initial max pooling layer to $(s/6)\times2$.

In action classification, we perform SoI pooling (i.e. 1D RoI pooling) to
extract a fixed size feature map for each obtained proposal. We illustrate the
mechanism of SoI pooling with output size 7 in Fig.~\ref{fig:context-classify}
(top). Note that as in the original design of RoI
pooling~\cite{girshick:iccv2015,ren:nips2015}, pooling is applied to the region
strictly within the proposal, which includes no temporal contexts. We propose
to extend the input extent of SoI pooling. As shown in
Fig.~\ref{fig:context-classify} (bottom), for a proposal of size $s$, the
extent of our SoI pooling covers not only the proposal segment, but also the
two segments of size $s/2$ immediately before and after the proposal, similar
to the classification of anchors. After SoI pooling, we add one fully-connected
layer, followed by a final fully-connected layer, which classifies the action
and regresses the boundaries.

\vspace{-3mm}

\paragraph{Contributions
beyond~\cite{dai:iccv2017,gao:bmvc2017,gao:iccv2017,xu:iccv2017}} Xu et
al.~\cite{xu:iccv2017} did not exploit any context features in either proposal
generation or action classification. Dai et al.~\cite{dai:iccv2017} included
context features when generating proposals, but used only the features within
the proposal in action classification. Gao et al. exploited context features in
either proposal generation only~\cite{gao:iccv2017} or both
stages~\cite{gao:bmvc2017}. However, they average-pooled the features within
the context regions, while we use temporal convolutions and SoI pooling to
encode the temporal structure of the features.

\subsection{Late Feature Fusion}
\label{sec:fusion}

\begin{figure}[t]
 \centering
 \includegraphics[width=0.96\linewidth]{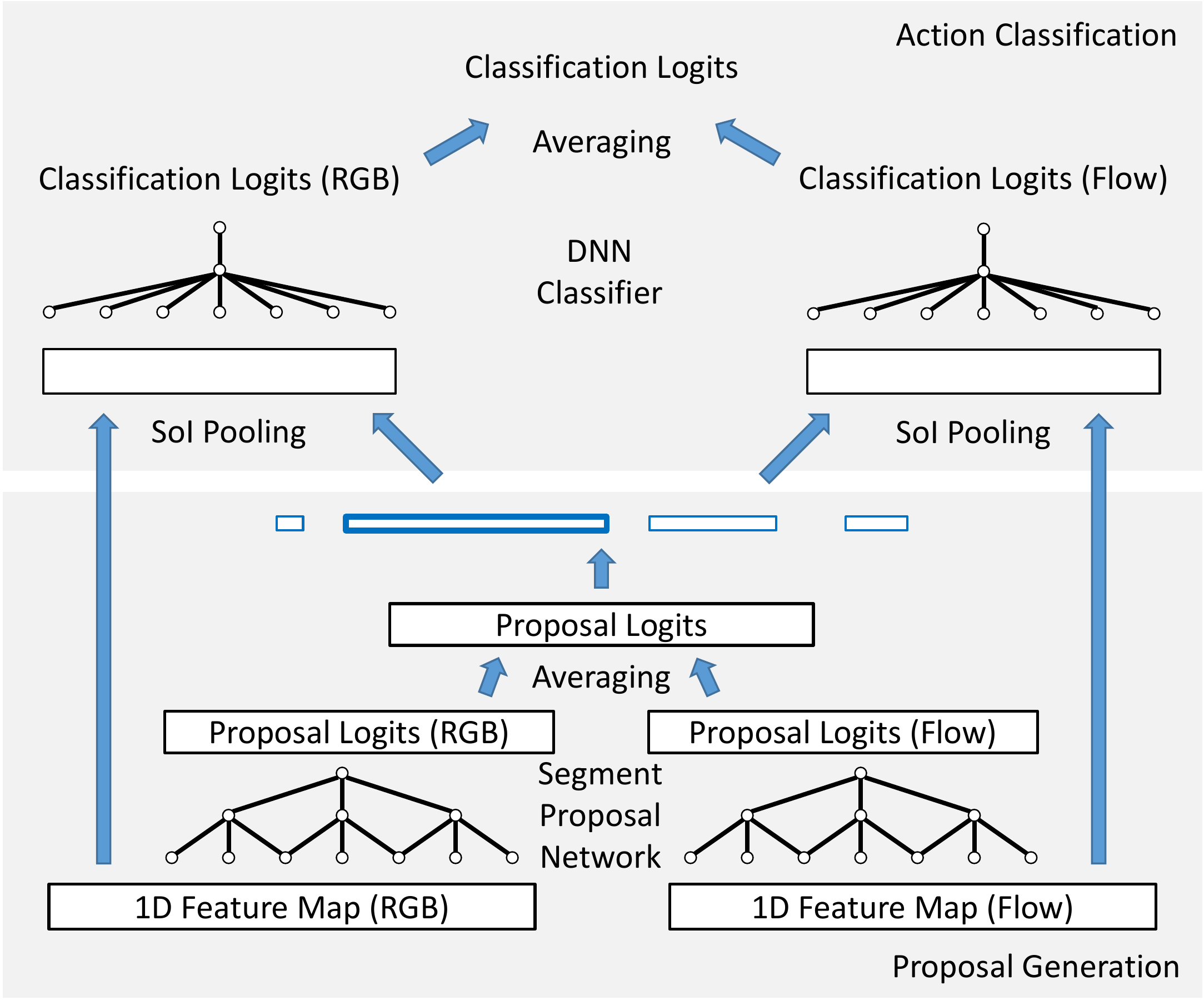}
 \vspace{-1mm} 
 \caption{\small The late fusion scheme for the two-stream Faster R-CNN
framework.}
 \vspace{-2mm} 
 \label{fig:late_fusion}
\end{figure}

In action classification, most of the state-of-the-art
methods~\cite{simonyan:nips2014,ng:cvpr2015,wang:eccv2016,carreira:cvpr2017,feichtenhofer:cvpr2017}
rely on a two-stream architecture, which parallelly processes two types of
input---RGB frames and pre-computed optical flow---and later fuses their
features to generate the final classification scores. We hypothesize such
two-stream input and feature fusion may also play an important role in temporal
action localization. Therefore we propose a \textit{late fusion} scheme for the
two-stream Faster R-CNN framework. Conceptually, this is equivalent to
performing the conventional late fusion in both the proposal generation and
action classification stage (Fig.~\ref{fig:late_fusion}). We first extract two
1D feature maps from RGB frames and stacked optical flow, respectively, using
two different networks. We process each feature map by a distinct Segment
Proposal Network, which parallelly generates the logits for anchor
classification and boundary regression. We use the element-wise average of the
logits from the two networks as the final logits to generate proposals. For
each proposal, we perform SoI pooling parallelly on both feature maps, and
apply a distinct DNN classifier on each output. Finally, the logits for action
classification and boundary regression from both DNN classifiers are
element-wisely averaged to generate the final detection output.

Note that a more straightforward way to fuse two features is through an
\textit{early fusion} scheme: we concatenate the two 1D feature maps in the
feature dimension, and apply the same pipeline as before
(Sec.~\ref{sec:receptive} and~\ref{sec:context}). We show by experiments that
the aforementioned late fusion scheme outperforms the early fusion scheme.

\vspace{-3mm}

\paragraph{Contributions
beyond~\cite{dai:iccv2017,gao:bmvc2017,gao:iccv2017,xu:iccv2017}} Xu et
al.~\cite{xu:iccv2017} only used a single-stream feature (C3D). Both Dai et al.
and Gao et al. used two-stream features, but either did not perform fusion
~\cite{gao:iccv2017} or only tried the early fusion
scheme~\cite{dai:iccv2017,gao:bmvc2017}.

\section{Experiments}

\paragraph{Dataset} We perform ablation studies and state-of-the-art
comparisons on the temporal action detection benchmark of
THUMOS'14~\cite{THUMOS14}. The dataset contains videos from 20 sports action
classes. Since the training set contains only trimmed videos with no temporal
annotations, we use the 200 untrimmed videos (3,007 action instances) in the
validation set to train our model. The test set consists of 213 videos (3,358
action instances). Each video is on average more than 3 minutes long, and
contains on average more than 15 action instances, making the task particularly
challenging. Besides THUMOS'14, we separately report our results on ActivityNet
v1.3~\cite{caba_heilbron:cvpr2015} at the end of the section.

\vspace{-3mm}

\paragraph{Evaluation Metrics} We consider two tasks: \textit{action proposal}
and \textit{action localization}. For action proposal, we calculate Average
Recall (AR) at different Average Number of Proposals per Video (AN) using the
public code provided by~\cite{escorcia:eccv2016}. AR is defined by the average
of all recall values using tIoU thresholds from 0.5 to 1 with a step size of
0.05. For action localization, we report mean Average Precision (mAP) using
different tIoU thresholds.

\vspace{-3mm}

\paragraph{Features} To extract the feature maps, we first train a two-stream
"Inflated 3D ConvNet" (I3D) model~\cite{carreira:cvpr2017} on the Kinetics
action classification dataset~\cite{kay:arxiv2017}. The I3D model builds upon
state-of-the-art image classification architectures (i.e.
Inception-v1~\cite{szegedy:cvpr2015}), but inflates their filters and pooling
kernels into 3D, leading to very deep, naturally spatiotemporal classifiers.
The model takes as input a stack of $64$ RGB/optical flow frames, performs
spatio-temporal convolutions, and extracts a $1024$-dimensional feature as the
output of an average pooling layer. We extract both RGB and optical flow frames
at $10$ frames per second (fps) as input to the I3D model. To compute optical
flow, we use a FlowNet~\cite{dosovitskiy:iccv2015} model trained on
artificially generated data followed by fine-tuning on the Kinetics dataset
using an unsupervised loss~\cite{vijayanarasimhan:arxiv2017}. After training on
Kinetics we fix the model and extract the $1024$-dimensional output of the
average pooling layer by stacking every $16$ RGB/optical flow frames in the
frame sequence. The input to our action localization model is thus two
$1024$-dimensional feature maps---for RGB and optical flow---sampled at $0.625$
fps from the input videos. 

\vspace{-3mm}

\paragraph{Implementation Details} Our implementation is based on the
TensorFlow Object Detection API~\cite{huang:cvpr2017}. In proposal generation,
we apply anchors of the following scales: $\{1, 2, 3, 4, 5, 6, 8, 11, 16\}$,
i.e. $K=9$. We set the number of filters to 256 for all convolutional and
fully-connected layers in the Segment Proposal Network and the DNN classifier.
We add a convolutional layer with kernel size 1 to reduce the feature dimension
to 256 before the Segment Proposal Network and after the SoI pooling layer. We
apply Non-Maximum Suppression (NMS) with tIoU threshold 0.7 on the proposal
output and keep the top 300 proposals for action classification. The same NMS
is applied to the final detection output for each action class separately. The
training of TAL-Net largely follows the Faster R-CNN implementation
in~\cite{huang:cvpr2017}. We provide the details in the supplementary material.


\vspace{-3mm}

\paragraph{Receptive Field Alignment} We validate the design for receptive
field alignment by comparing four baselines: (1) a single-tower network with no
temporal convolutions (Single), where each anchor is classified solely based on
the feature at its center location; (2) a single-tower network with non-dilated
temporal convolutions (Single+TConv), which represents the default Faster R-CNN
architecture; (3) a multi-tower network with non-dilated temporal convolutions
(Multi+TConv); (4) a multi-tower network with dilated temporal convolutions
(Multi+Dilated, the proposed architecture). All temporal ConvNets have two
layers, both with kernel size 3. Here we consider only a single-steam feature
(i.e. RGB or flow) and evaluate the generated proposal with AR-AN. The results
are reported in Tab.~\ref{tab:receptive} (top for RGB and bottom for flow). The
trend is consistent on both features: Single performs the worst, since it
relies only on the context at the center location; Single+TConv and Multi+TConv
both perform better than Single, but still, suffer from irrelevant context due
to misaligned receptive fields; Multi-Dilated outperforms the others, as the
receptive fields are properly aligned with the span of anchors.

\vspace{-3mm}

\paragraph{Context Feature Extraction} We first validate our design for context
feature extraction in proposal generation. Tab.~\ref{tab:context-proposal}
compares the generated proposals before and after incorporating context
features (top for RGB and bottom for flow). We achieve higher AR on both
streams after the context features are included. Next, given better proposals,
we evaluate context feature extraction in action classification.
Tab.~\ref{tab:context-localize} compares the action localization results before
and after incorporating context features (top for RGB and bottom for flow).
Similarly, we achieve higher mAP nearly at all AN values on both streams after
including the context features.

\begin{table}[t]
 \centering
 \small
 \begin{minipage}{0.48\textwidth}
  \centering
  \begin{tabular}{l||C{0.62cm}C{0.62cm}C{0.62cm}C{0.62cm}C{0.62cm}}
   \hline \TBstrut
   AN              & 10            & 20            & 50            & 100           & 200           \\
   \hline \Tstrut
   Single          & ~~9.4         & 15.3          & 25.3          & 33.9          & 41.3          \\
   Single + TConv  & 12.9          & 20.0          & 30.3          & 37.6          & 44.0          \\
   Multi + TConv   & 13.4          & 20.6          & 31.1          & 38.1          & 43.7          \\ \Bstrut
   Multi + Dilated & \textbf{14.0} & \textbf{21.7} & \textbf{31.9} & \textbf{38.8} & \textbf{44.7} \\
   \hline
   \hline \Tstrut
   Single          & 11.0          & 18.0          & 28.9          & 36.8          & 43.6          \\
   Single + TConv  & 15.1          & 23.2          & 33.7          & 40.0          & 44.7          \\
   Multi + TConv   & 15.7          & 24.0          & 35.0          & 41.1          & 46.2          \\ \Bstrut
   Multi + Dilated & \textbf{16.3} & \textbf{25.4} & \textbf{35.8} & \textbf{42.3} & \textbf{47.5} \\
   \hline
  \end{tabular}
 \end{minipage}
 \vspace{-2mm}
 \caption{\small Results for receptive field alignment on proposal generation
in AR (\%). Top: RGB stream. Bottom: Flow stream.}
 \vspace{-1mm}
 \label{tab:receptive}
\end{table}

\begin{table}[t]
 \centering
 \small
 \setlength{\tabcolsep}{3.47pt}
 \begin{tabular}{l||C{0.62cm}C{0.62cm}C{0.62cm}C{0.62cm}C{0.62cm}}
  \hline \TBstrut
  AN                         & 10            & 20            & 50            & 100           & 200           \\
  \hline \Tstrut
  Multi + Dilated            & 14.0          & 21.7          & 31.9          & 38.8          & 44.7          \\ \Bstrut
  Multi + Dilated + Context  & \textbf{15.1} & \textbf{22.2} & \textbf{32.3} & \textbf{39.9} & \textbf{46.8} \\
  \hline
  \hline \Tstrut
  Multi + Dilated            & 16.3          & 25.4          & 35.8          & 42.3          & 47.5          \\ \Bstrut
  Multi + Dilated + Context  & \textbf{17.4} & \textbf{26.5} & \textbf{36.5} & \textbf{43.3} & \textbf{48.6} \\
  \hline
 \end{tabular}
 \vspace{-2mm}
 \caption{\small Results for incorporating context features in proposal
generation in AR (\%). Top: RGB stream. Bottom: Flow stream.}
 \vspace{-1mm}
 \label{tab:context-proposal}
\end{table}

\begin{table}[t]
 \centering
 \small
 \setlength{\tabcolsep}{3.47pt}
 \begin{tabular}{l||C{0.724cm}C{0.724cm}C{0.724cm}C{0.724cm}C{0.724cm}}
  \hline \TBstrut
  tIoU                   & 0.1           & 0.3           & 0.5           & 0.7           & 0.9           \\
  \hline \Tstrut
  SoI Pooling            & 44.9          & 38.4          & 28.5          & 13.0          & \textbf{0.6}  \\ \Bstrut
  SoI Pooling + Context  & \textbf{49.3} & \textbf{42.6} & \textbf{31.9} & \textbf{14.2} & \textbf{0.6}  \\
  \hline
  \hline \Tstrut
  SoI Pooling            & 49.8          & 45.7          & 37.4          & \textbf{18.8} & 0.7           \\ \Bstrut
  SoI Pooling + Context  & \textbf{54.3} & \textbf{48.8} & \textbf{38.2} & 18.6          & \textbf{0.9}  \\
  \hline
 \end{tabular}
 \vspace{-2mm}
 \caption{\small Results for incorporating context features in action
classification in mAP (\%). Top: RGB stream. Bottom: Flow stream.}
 \vspace{-1mm}
 \label{tab:context-localize}
\end{table}

\begin{table}[t]
 \centering
 \small
 \begin{tabular}{l||ccccc}
  \hline \TBstrut
  tIoU            & 0.1           & 0.3           & 0.5           & 0.7           & 0.9           \\
  \hline \Tstrut
  RGB             & 49.3          & 42.6          & 31.9          & 14.2          & 0.6           \\ \Bstrut
  Flow            & 54.3          & 48.8          & 38.2          & 18.6          & \textbf{0.9}  \\ \hline \Tstrut
  Early Fusion    & \textbf{60.5} & 52.8          & 40.8          & 19.3          & 0.8           \\ \Bstrut
  Late Fusion     & 59.8          & \textbf{53.2} & \textbf{42.8} & \textbf{20.8} & \textbf{0.9}  \\
  \hline
 \end{tabular}
 \vspace{-2mm}
 \caption{\small Results for late feature fusion in mAP (\%).}
 \vspace{-2mm} 
 \label{tab:late_fusion}
\end{table}

\vspace{-3mm}

\paragraph{Late Feature Fusion} Tab.~\ref{tab:late_fusion} reports the action
localization results of the two single-stream networks and the early and late
fusion schemes. First, the flow based feature outperforms the RGB based
feature, which coheres with the common observations in action
classification~\cite{simonyan:nips2014,wang:eccv2016,carreira:cvpr2017,feichtenhofer:cvpr2017}.
Second, the fused features outperform the two single-stream features,
suggesting the RGB and flow features complement each other. Finally, the late
fusion scheme outperforms the early fusion scheme except at tIoU threshold 0.1,
validating our proposed design.

\begin{figure}[t]
 \centering
 \includegraphics[width=0.70\linewidth]{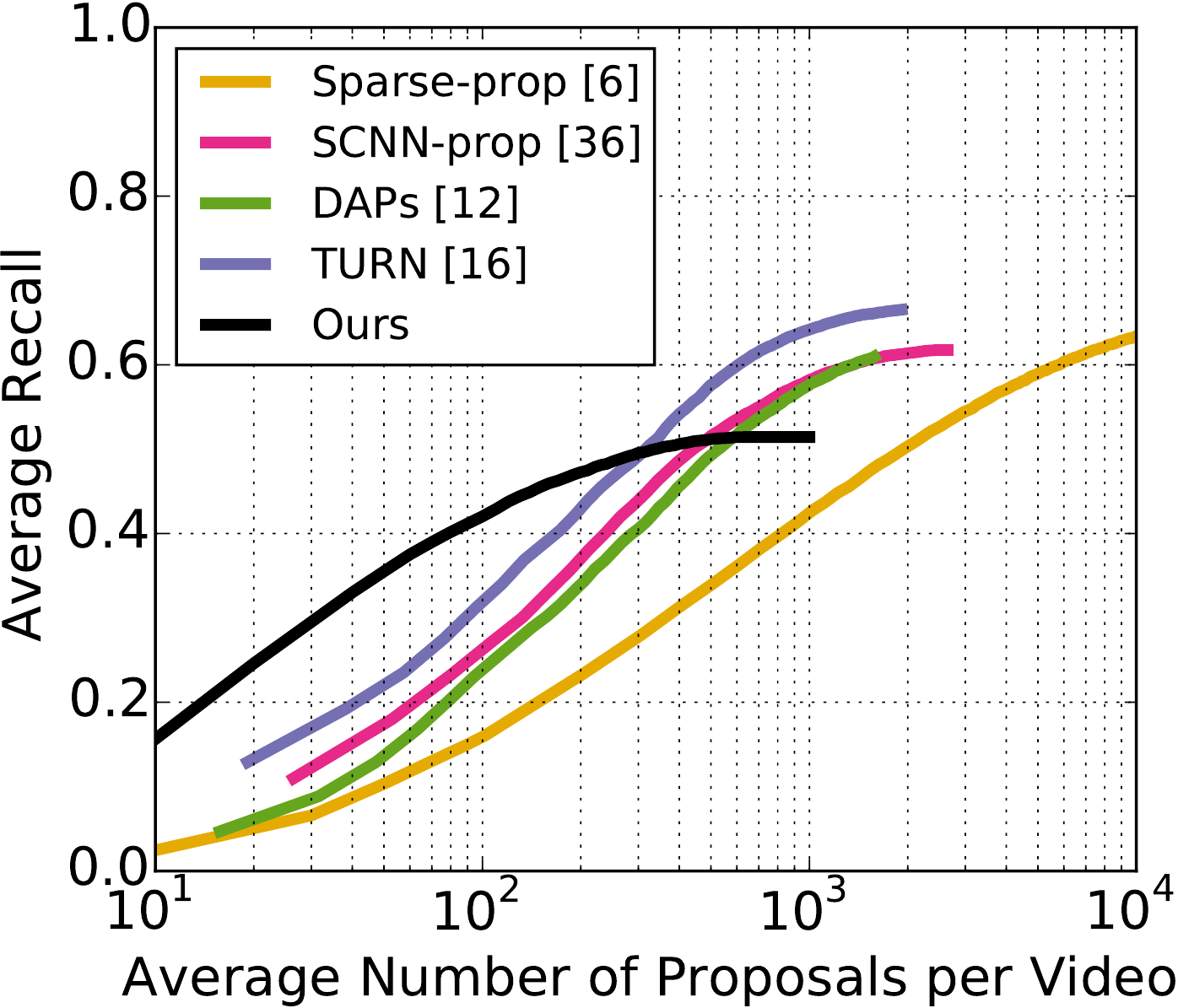}
 \vspace{-1mm}
 \caption{\small Our action proposal result in AR-AN (\%) on THUMOS'14
comparing with other state-of-the-art methods.}
 \vspace{-2mm}
 \label{fig:benchmark-proposal}
\end{figure}

\begin{table}[t]
 \centering
 \footnotesize
 \setlength{\tabcolsep}{3.96pt}
 \setlength\extrarowheight{1.5pt}
 \begin{tabular}{lccccccc}
  \hline \TBstrut
  tIoU                                                & 0.1           & 0.2           & 0.3           & 0.4           & 0.5           & 0.6           & 0.7           \\
  \hline \Tstrut
  Karaman et al.~\cite{karaman:2014}                  & ~~4.6         & ~~3.4         & ~~2.4         & ~~1.4         & ~~0.9         & --            & --            \\
  Oneata et al.~\cite{oneata:2014}                    & 36.6          & 33.6          & 27.0          & 20.8          & 14.4          & --            & --            \\
  Wang et al.~\cite{wang:2014}                        & 18.2          & 17.0          & 14.0          & 11.7          & ~~8.3         & --            & --            \\
  Caba Heilbron et al.~\cite{caba_heilbron:cvpr2016}  & --            & --            & --            & --            & 13.5          & --            & --            \\ 
  Richard and Gall~\cite{richard:cvpr2016}            & 39.7          & 35.7          & 30.0          & 23.2          & 15.2          & --            & --            \\
  Shou et al.~\cite{shou:cvpr2016}                    & 47.7          & 43.5          & 36.3          & 28.7          & 19.0          & 10.3          & ~~5.3         \\
  Yeung et al.~\cite{yeung:cvpr2016}                  & 48.9          & 44.0          & 36.0          & 26.4          & 17.1          & --            & --            \\
  Yuan et al.~\cite{yuan:cvpr2016}                    & 51.4          & 42.6          & 33.6          & 26.1          & 18.8          & --            & --            \\
  Escorcia et al.~\cite{escorcia:eccv2016}            & --            & --            & --            & --            & 13.9          & --            & --            \\
  Buch et al.~\cite{buch:cvpr2017}                    & --            & --            & 37.8          & --            & 23.0          & --            & --            \\
  Shou et al.~\cite{shou:cvpr2017}                    & --            & --            & 40.1          & 29.4          & 23.3          & 13.1          & ~~7.9         \\
  Yuan et al.~\cite{yuan:cvpr2017}                    & 51.0          & 45.2          & 36.5          & 27.8          & 17.8          & --            & --            \\
  Buch et al.~\cite{buch:bmvc2017}                    & --            & --            & 45.7          & --            & 29.2          & --            & ~~9.6         \\
  Gao et al.~\cite{gao:bmvc2017}                      & 60.1          & 56.7          & 50.1          & 41.3          & 31.0          & 19.1          & ~~9.9         \\
  Hou et al.~\cite{hou:bmvc2017}                      & 51.3          & --            & 43.7          & --            & 22.0          & --            & --            \\
  Dai et al.~\cite{dai:iccv2017}                      & --            & --            & --            & 33.3          & 25.6          & 15.9          & ~~9.0         \\
  Gao et al.~\cite{gao:iccv2017}                      & 54.0          & 50.9          & 44.1          & 34.9          & 25.6          & --            & --            \\
  Xu et al.~\cite{xu:iccv2017}                        & 54.5          & 51.5          & 44.8          & 35.6          & 28.9          & --            & --            \\ \Bstrut
  Zhao et al.~\cite{zhao:iccv2017}                    & \textbf{66.0} & \textbf{59.4} & 51.9          & 41.0          & 29.8          & --            & --            \\ \hline
  \TBstrut
  Ours                                                & 59.8          & 57.1          & \textbf{53.2} & \textbf{48.5} & \textbf{42.8} & \textbf{33.8} & \textbf{20.8} \\
  \hline
 \end{tabular}
 \vspace{-2mm}
 \caption{\small Action localization mAP (\%) on THUMOS'14.}
 \vspace{-2mm}
 \label{tab:benchmark-localize}
\end{table}

\begin{figure*}[t]
 \centering
 \begin{minipage}{0.058\textwidth} \centering \includegraphics[width=1.00\textwidth]{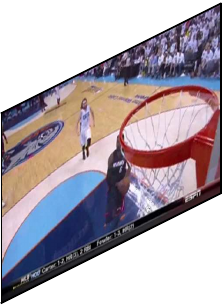} \end{minipage}
 \begin{minipage}{0.058\textwidth} \centering \includegraphics[width=1.00\textwidth]{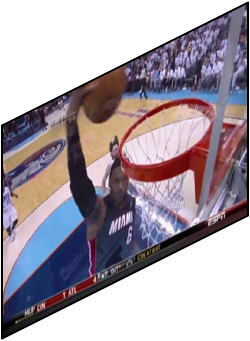} \end{minipage}
 \begin{minipage}{0.058\textwidth} \centering \includegraphics[width=1.00\textwidth]{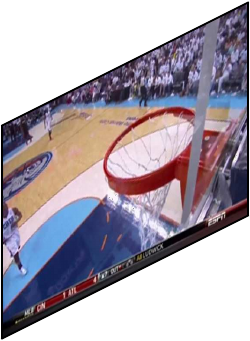} \end{minipage}
 \begin{minipage}{0.058\textwidth} \centering \includegraphics[width=1.00\textwidth]{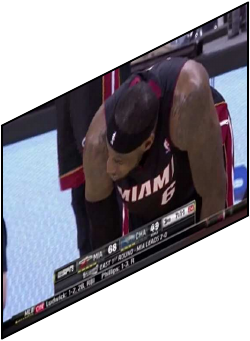} \end{minipage}
 \begin{minipage}{0.058\textwidth} \centering \includegraphics[width=1.00\textwidth]{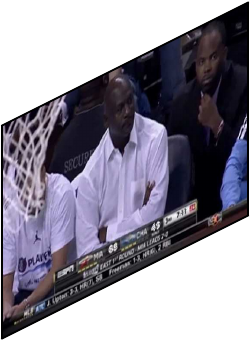} \end{minipage}
 \begin{minipage}{0.058\textwidth} \centering \includegraphics[width=1.00\textwidth]{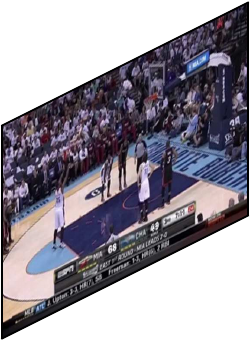} \end{minipage}
 \begin{minipage}{0.058\textwidth} \centering \includegraphics[width=1.00\textwidth]{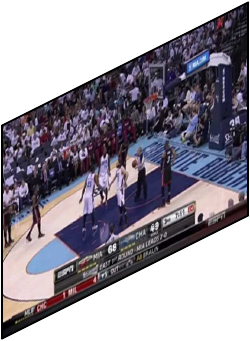} \end{minipage}
 \begin{minipage}{0.058\textwidth} \centering \includegraphics[width=1.00\textwidth]{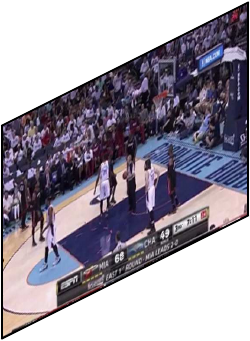} \end{minipage}
 \begin{minipage}{0.058\textwidth} \centering \includegraphics[width=1.00\textwidth]{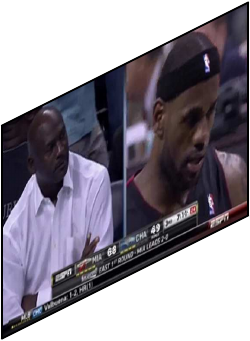} \end{minipage}
 \begin{minipage}{0.058\textwidth} \centering \includegraphics[width=1.00\textwidth]{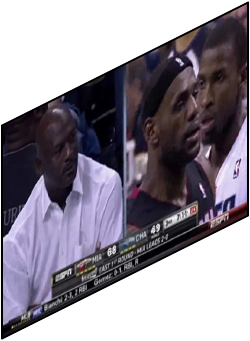} \end{minipage}
 \begin{minipage}{0.058\textwidth} \centering \includegraphics[width=1.00\textwidth]{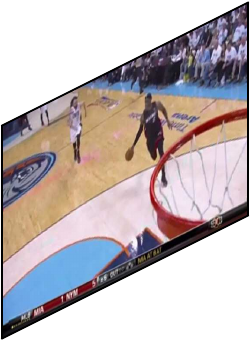} \end{minipage}
 \begin{minipage}{0.058\textwidth} \centering \includegraphics[width=1.00\textwidth]{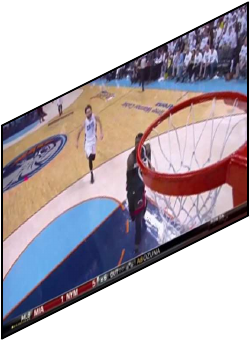} \end{minipage}
 \begin{minipage}{0.058\textwidth} \centering \includegraphics[width=1.00\textwidth]{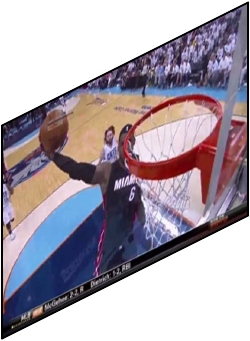} \end{minipage}
 \begin{minipage}{0.058\textwidth} \centering \includegraphics[width=1.00\textwidth]{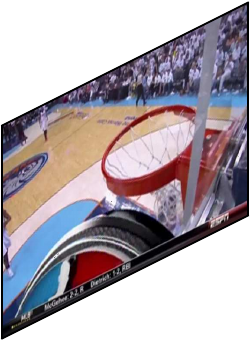} \end{minipage}
 \begin{minipage}{0.058\textwidth} \centering \includegraphics[width=1.00\textwidth]{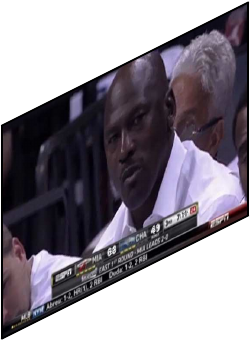} \end{minipage}
 \begin{minipage}{0.058\textwidth} \centering \includegraphics[width=1.00\textwidth]{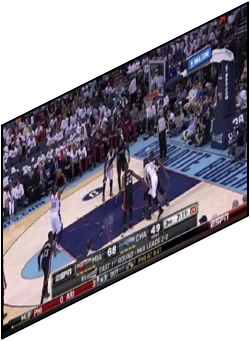} \end{minipage}
 \\ \vspace{2mm}
 \includegraphics[width=1.00\textwidth]{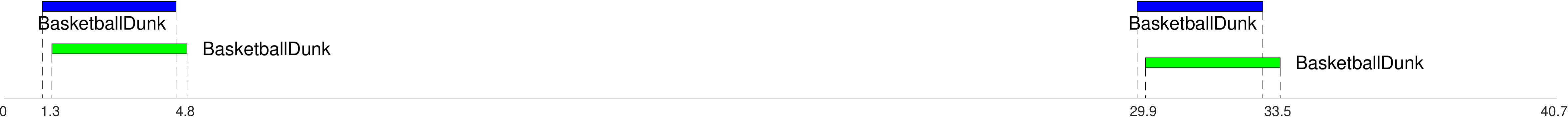}
 \\ \vspace{2mm}
 \begin{minipage}{0.058\textwidth} \centering \includegraphics[width=1.00\textwidth]{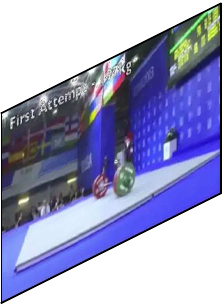} \end{minipage}
 \begin{minipage}{0.058\textwidth} \centering \includegraphics[width=1.00\textwidth]{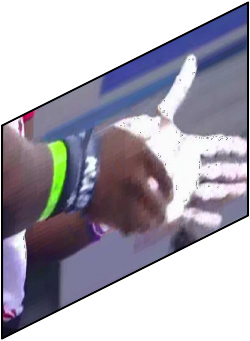} \end{minipage}
 \begin{minipage}{0.058\textwidth} \centering \includegraphics[width=1.00\textwidth]{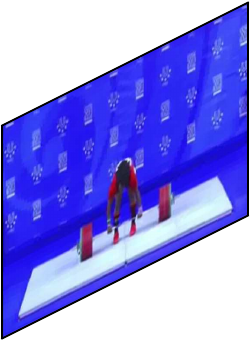} \end{minipage}
 \begin{minipage}{0.058\textwidth} \centering \includegraphics[width=1.00\textwidth]{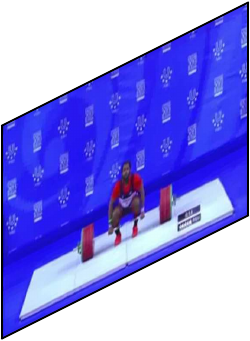} \end{minipage}
 \begin{minipage}{0.058\textwidth} \centering \includegraphics[width=1.00\textwidth]{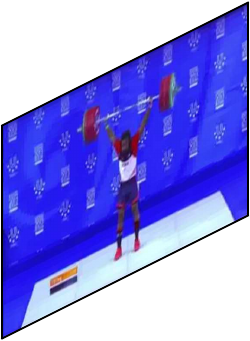} \end{minipage}
 \begin{minipage}{0.058\textwidth} \centering \includegraphics[width=1.00\textwidth]{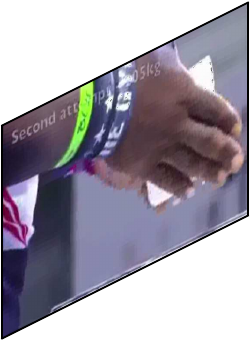} \end{minipage}
 \begin{minipage}{0.058\textwidth} \centering \includegraphics[width=1.00\textwidth]{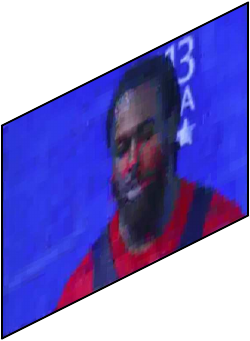} \end{minipage}
 \begin{minipage}{0.058\textwidth} \centering \includegraphics[width=1.00\textwidth]{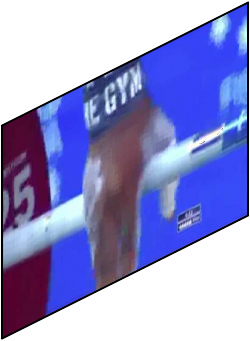} \end{minipage}
 \begin{minipage}{0.058\textwidth} \centering \includegraphics[width=1.00\textwidth]{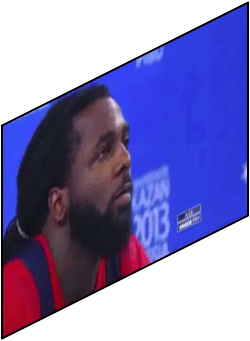} \end{minipage}
 \begin{minipage}{0.058\textwidth} \centering \includegraphics[width=1.00\textwidth]{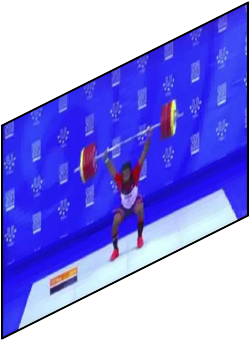} \end{minipage}
 \begin{minipage}{0.058\textwidth} \centering \includegraphics[width=1.00\textwidth]{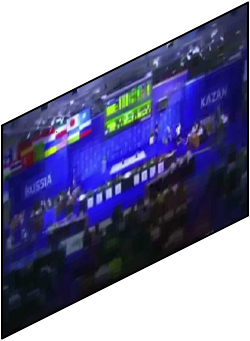} \end{minipage}
 \begin{minipage}{0.058\textwidth} \centering \includegraphics[width=1.00\textwidth]{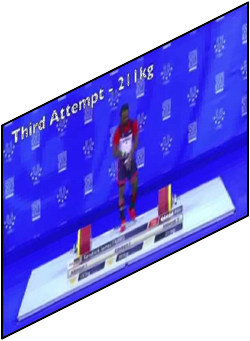} \end{minipage}
 \begin{minipage}{0.058\textwidth} \centering \includegraphics[width=1.00\textwidth]{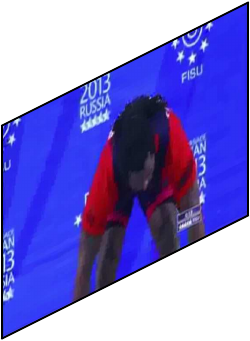} \end{minipage}
 \begin{minipage}{0.058\textwidth} \centering \includegraphics[width=1.00\textwidth]{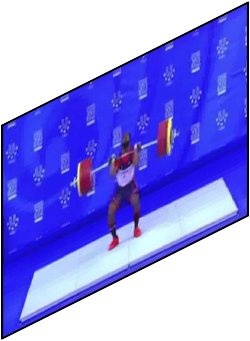} \end{minipage}
 \begin{minipage}{0.058\textwidth} \centering \includegraphics[width=1.00\textwidth]{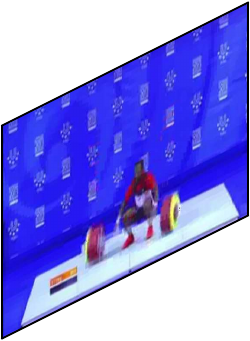} \end{minipage}
 \begin{minipage}{0.058\textwidth} \centering \includegraphics[width=1.00\textwidth]{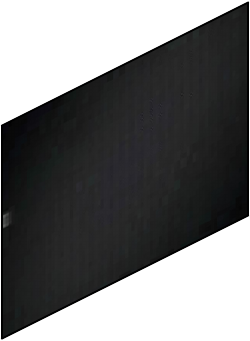} \end{minipage}
 \\ \vspace{2mm}
 \includegraphics[width=1.00\textwidth]{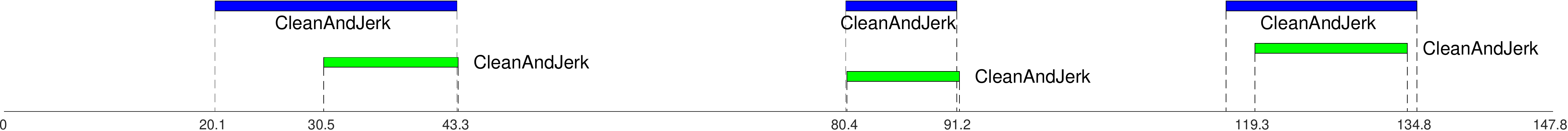}
 \\ \vspace{2mm}
 \begin{minipage}{0.058\textwidth} \centering \includegraphics[width=1.00\textwidth]{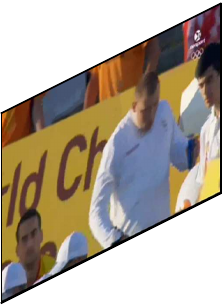} \end{minipage}
 \begin{minipage}{0.058\textwidth} \centering \includegraphics[width=1.00\textwidth]{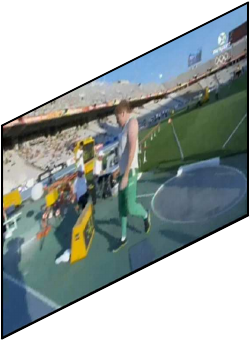} \end{minipage}
 \begin{minipage}{0.058\textwidth} \centering \includegraphics[width=1.00\textwidth]{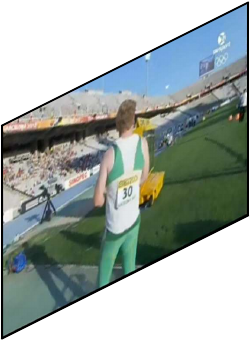} \end{minipage}
 \begin{minipage}{0.058\textwidth} \centering \includegraphics[width=1.00\textwidth]{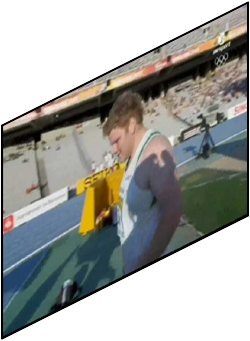} \end{minipage}
 \begin{minipage}{0.058\textwidth} \centering \includegraphics[width=1.00\textwidth]{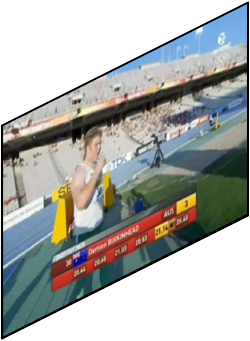} \end{minipage}
 \begin{minipage}{0.058\textwidth} \centering \includegraphics[width=1.00\textwidth]{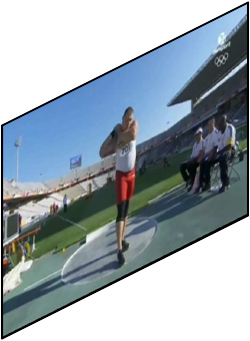} \end{minipage}
 \begin{minipage}{0.058\textwidth} \centering \includegraphics[width=1.00\textwidth]{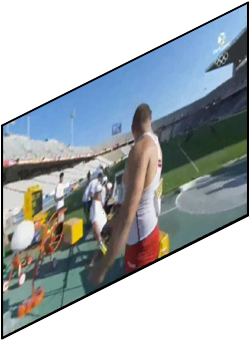} \end{minipage}
 \begin{minipage}{0.058\textwidth} \centering \includegraphics[width=1.00\textwidth]{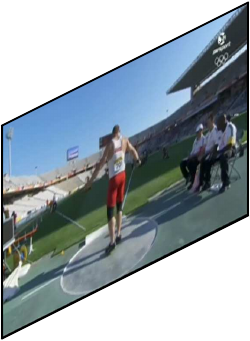} \end{minipage}
 \begin{minipage}{0.058\textwidth} \centering \includegraphics[width=1.00\textwidth]{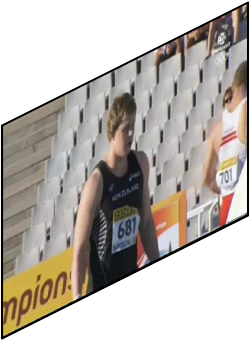} \end{minipage}
 \begin{minipage}{0.058\textwidth} \centering \includegraphics[width=1.00\textwidth]{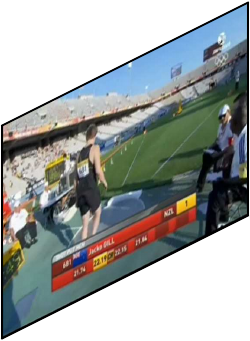} \end{minipage}
 \begin{minipage}{0.058\textwidth} \centering \includegraphics[width=1.00\textwidth]{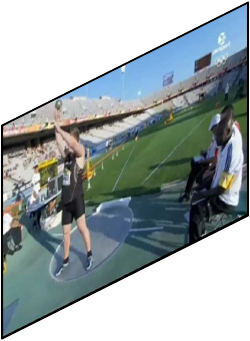} \end{minipage}
 \begin{minipage}{0.058\textwidth} \centering \includegraphics[width=1.00\textwidth]{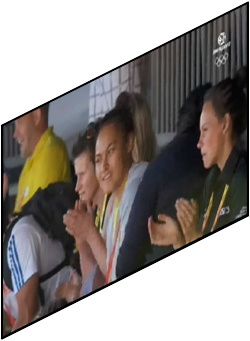} \end{minipage}
 \begin{minipage}{0.058\textwidth} \centering \includegraphics[width=1.00\textwidth]{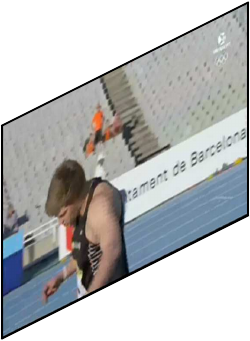} \end{minipage}
 \begin{minipage}{0.058\textwidth} \centering \includegraphics[width=1.00\textwidth]{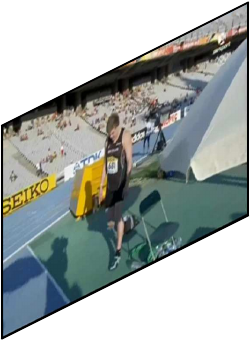} \end{minipage}
 \begin{minipage}{0.058\textwidth} \centering \includegraphics[width=1.00\textwidth]{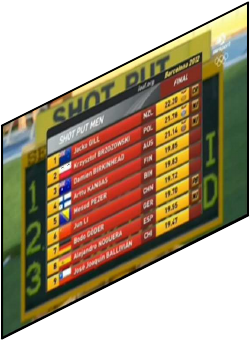} \end{minipage}
 \begin{minipage}{0.058\textwidth} \centering \includegraphics[width=1.00\textwidth]{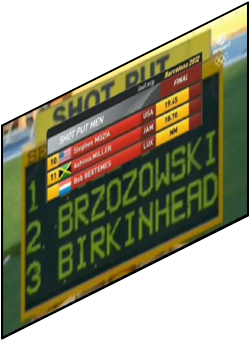} \end{minipage}
 \\ \vspace{2mm}
 \includegraphics[width=1.00\textwidth]{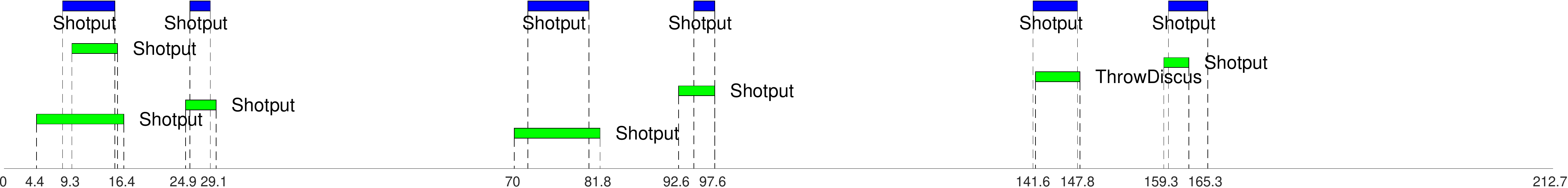}
 \caption{\small Qualitative examples of the top localized actions on
THUMOS'14. Each consists of a sequence of frames sampled from a full test
video, the ground-truth (blue) and predicted (green) action segments and class
labels, and a temporal axis showing the time in seconds.}
 \label{fig:qualitative}
\end{figure*}

\vspace{-3mm}


\paragraph{State-of-the-Art Comparisons} We compare TAL-Net with
state-of-the-art methods on both action proposal and localization.
Fig.~\ref{fig:benchmark-proposal} shows the AR-AN curves for action proposal.
TAL-Net outperforms all other methods in the low AN region, suggesting our top
proposals have higher quality. Although our AR saturates earlier as AN
increases, this is because we extract features at a much lower frequency (i.e.
0.625 fps) due to the high computational demand of the I3D models. This reduces
the density of anchors and lowers the upper bound of the recall.
Tab.~\ref{tab:benchmark-localize} compares the mAP for action localization.
TAL-Net achieves the highest mAP when the tIoU threshold is greater than 0.2,
suggesting it can localize the boundaries more accurately. We particularly
highlight our result at tIoU threshold 0.5, where TAL-Net outperforms the
state-of-the-art by 11.8\% mAP (42.8\% versus 31.0\% from Gao et
al.~\cite{gao:bmvc2017}).

\vspace{-3mm}

\paragraph{Qualitative Results} Fig.~\ref{fig:qualitative} shows 
qualitative examples of the top localized actions on THUMOS'14.
Each
consists of a sequence of frames sampled from a full test video, the
ground-truth (blue) and predicted (green) action segments and class labels, and
a temporal axis showing the time in seconds. In the top example, our method
accurately localizes both
instances in the video. In the middle example, the action classes are correctly
classified, but the start of the leftmost prediction is inaccurate, due to
subtle differences between preparation and the start of the action.
In the bottom,
``ThrowDiscus'' is misclassified due to similar context.

\vspace{-3mm}

\paragraph{Results on ActivityNet} Tab.~\ref{tab:activitynet} shows our action
localization results on the ActivityNet v1.3 validation set along with other
recent published results. 
TAL-Net outperforms other Faster R-CNN based methods at tIoU threshold 0.5
(38.23\% vs.\ 36.44\% from Dai et al.~\cite{dai:iccv2017} and 26.80\% from Xu
et al.~\cite{xu:iccv2017}). Note that THUMOS'14 is a better dataset for
evaluating action localization than ActivityNet, as the former has more action
instances per video and each video contains a larger portion of background
activity: on average, the THUMOS'14 training set has 15 instances per video and
each video has 71\% background, while the ActivityNet training set has only 1.5
instances per video and each video has only 36\% background.

\begin{table}[t]
 \centering
 \small
 \begin{tabular}{lcccc}
  \hline \TBstrut
  tIoU                                                & 0.5            & 0.75           & 0.95          & Average        \\
  \hline \Tstrut
  Singh and Cuzzolin~\cite{singh:2016}                & 34.47          & --             & --            & --             \\
  Wang and Tao~\cite{wang:2016}                       & 43.65          & --             & --            & --             \\
  Shou et al.~\cite{shou:cvpr2017}                    & \textbf{45.30} & \textbf{26.00} & 0.20          & \textbf{23.80} \\ 
  Dai et al.~\cite{dai:iccv2017}                      & 36.44          & 21.15          & \textbf{3.90} & --             \\ \Bstrut
  Xu et al.~\cite{xu:iccv2017}                        & 26.80          & --             & --            & 12.70          \\ \hline \TBstrut 
  Ours                                                & 38.23          & 18.30          & 1.30          & 20.22          \\
  \hline
 \end{tabular}
 \vspace{-2mm}
 \caption{\small Action localization mAP (\%) on ActivityNet v1.3 (val).}
 \vspace{-2mm}
 \label{tab:activitynet}
\end{table}


\section{Conclusion}
We introduce TAL-Net, an improved approach to temporal action localization in
video that is inspired by the Faster RCNN object detection framework. TAL-Net
features three novel architectural changes that address three key shortcomings
of existing approaches: (1) receptive field alignment; (2) context feature
extraction; and (3) late feature fusion. We achieve state-ofthe-art performance
for both action proposal and localization on THUMOS’14 detection benchmark and
competitive performance on ActivityNet challenge.

\vspace{-3mm}

\paragraph{Acknowledgement} We thank Jo{\~a}o Carreira and Susanna Ricco for
their help on the I3D models and optical flow.

{\small
\bibliographystyle{ieee}
\bibliography{egbib}

\begin{thebibliography}{10}\itemsep=-1pt

\bibitem{yjxiong:code}
\url{https://github.com/yjxiong/action-detection}.

\bibitem{buch:bmvc2017}
S.~Buch, V.~Escorcia, B.~Ghanem, L.~Fei-Fei, and J.~C. Niebles.
\newblock End-to-end, single-stream temporal action detection in untrimmed
  videos.
\newblock In {\em BMVC}, 2017.

\bibitem{buch:cvpr2017}
S.~Buch, V.~Escorcia, C.~Shen, B.~Ghanem, and J.~C. Niebles.
\newblock {SST}: Single-stream temporal action proposals.
\newblock In {\em CVPR}, 2017.

\bibitem{caba_heilbron:cvpr2017}
F.~{Caba Heilbron}, W.~Barrios, V.~Escorcia, and B.~Ghanem.
\newblock {SCC}: Semantic context cascade for efficient action detection.
\newblock In {\em CVPR}, 2017.

\bibitem{caba_heilbron:cvpr2015}
F.~{Caba Heilbron}, V.~Escorcia, B.~Ghanem, and J.~C. Niebles.
\newblock {A}ctivity{N}et: A large-scale video benchmark for human activity
  understanding.
\newblock In {\em CVPR}, 2015.

\bibitem{caba_heilbron:cvpr2016}
F.~{Caba Heilbron}, J.~C. Niebles, and B.~Ghanem.
\newblock Fast temporal activity proposals for efficient detection of human
  actions in untrimmed videos.
\newblock In {\em CVPR}, 2016.

\bibitem{carreira:cvpr2017}
J.~Carreira and A.~Zisserman.
\newblock Quo vadis, action recognition? a new model and the {Kinetics}
  dataset.
\newblock In {\em CVPR}, 2017.

\bibitem{chen:iclr2015}
L.-C. Chen, G.~Papandreou, I.~Kokkinos, K.~Murphy, and A.~L. Yuille.
\newblock Semantic image segmentation with deep convolutional nets and fully
  connected {CRF}s.
\newblock In {\em ICLR}, 2015.

\bibitem{dai:iccv2017}
X.~Dai, B.~Singh, G.~Zhang, L.~S. Davis, and Y.~Q. Chen.
\newblock Temporal context network for activity localization in videos.
\newblock In {\em ICCV}, 2017.

\bibitem{dave:cvpr2017}
A.~Dave, O.~Russakovsky, and D.~Ramanan.
\newblock Predictive-corrective networks for action detection.
\newblock In {\em CVPR}, 2017.

\bibitem{dosovitskiy:iccv2015}
A.~Dosovitskiy, P.~Fischer, E.~Ilg, P.~Hausser, C.~Hazırbas, V.~Golkov,
  P.~van~der Smagt, D.~Cremers, and T.~Brox.
\newblock {F}low{N}et: Learning optical flow with convolutional networks.
\newblock In {\em ICCV}, 2015.

\bibitem{escorcia:eccv2016}
V.~Escorcia, F.~{Caba Heilbron}, J.~C. Niebles, and B.~Ghanem.
\newblock {DAP}s: Deep action proposals for action understanding.
\newblock In {\em ECCV}, 2016.

\bibitem{everingham:ijcv2015}
M.~Everingham, S.~M.~A. Eslami, L.~{Van Gool}, C.~K.~I. Williams, J.~Winn, and
  A.~Zisserman.
\newblock The {PASCAL} visual object classes challenge: A retrospective.
\newblock {\em IJCV}, 111(1):98--136, Jan 2015.

\bibitem{feichtenhofer:cvpr2017}
C.~Feichtenhofer, A.~Pinz, and R.~P. Wildes.
\newblock Spatiotemporal multiplier networks for video action recognition.
\newblock In {\em CVPR}, 2017.

\bibitem{gao:bmvc2017}
J.~Gao, Z.~Yang, and R.~Nevatia.
\newblock Cascaded boundary regression for temporal action detection.
\newblock In {\em BMVC}, 2017.

\bibitem{gao:iccv2017}
J.~Gao, Z.~Yang, C.~Sun, K.~Chen, and R.~Nevatia.
\newblock {TURN} {TAP}: Temporal unit regression network for temporal action
  proposals.
\newblock In {\em ICCV}, 2017.

\bibitem{girshick:iccv2015}
R.~Girshick.
\newblock Fast {R}-{CNN}.
\newblock In {\em ICCV}, 2015.

\bibitem{girshick:cvpr2014}
R.~Girshick, J.~Donahue, T.~Darrell, and J.~Malik.
\newblock Rich feature hierarchies for accurate object detection and semantic
  segmentation.
\newblock In {\em CVPR}, 2014.

\bibitem{gkioxari:cvpr2015}
G.~Gkioxari and J.~Malik.
\newblock Finding action tubes.
\newblock In {\em CVPR}, 2015.

\bibitem{hou:bmvc2017}
R.~Hou, R.~Sukthankar, and M.~Shah.
\newblock Real-time temporal action localization in untrimmed videos by
  sub-action discovery.
\newblock In {\em BMVC}, 2017.

\bibitem{huang:cvpr2017}
J.~Huang, V.~Rathod, C.~Sun, M.~Zhu, A.~Korattikara, A.~Fathi, I.~Fischer,
  Z.~Wojna, Y.~Song, S.~Guadarrama, and K.~Murphy.
\newblock Speed/accuracy trade-offs for modern convolutional object detectors.
\newblock In {\em CVPR}, 2017.

\bibitem{THUMOS14}
Y.-G. Jiang, J.~Liu, A.~Roshan~Zamir, G.~Toderici, I.~Laptev, M.~Shah, and
  R.~Sukthankar.
\newblock {THUMOS} challenge: Action recognition with a large number of
  classes.
\newblock \url{http://crcv.ucf.edu/THUMOS14/}, 2014.

\bibitem{kalogeiton:iccv2017}
V.~Kalogeiton, P.~Weinzaepfel, V.~Ferrari, and C.~Schmid.
\newblock Action tubelet detector for spatio-temporal action localization.
\newblock In {\em ICCV}, 2017.

\bibitem{karaman:2014}
S.~Karaman, L.~Seidenari, and A.~D. Bimbo.
\newblock Fast saliency based pooling of fisher encoded dense trajectories.
\newblock \url{http://crcv.ucf.edu/THUMOS14/}, 2014.

\bibitem{kay:arxiv2017}
W.~Kay, J.~Carreira, K.~Simonyan, B.~Zhang, C.~Hillier, S.~Vijayanarasimhan,
  F.~Viola, T.~Green, T.~Back, P.~Natsev, M.~Suleyman, and A.~Zisserman.
\newblock The {K}inetics human action video dataset.
\newblock {\em arXiv preprint arXiv:1705.06950}, 2017.

\bibitem{lea:cvpr2017}
C.~Lea, M.~Flynn, R.~Vidal, A.~Reiter, and G.~Hager.
\newblock Temporal convolutional networks for action segmentation and
  detection.
\newblock In {\em CVPR}, 2017.

\bibitem{lin:acmmm2017}
T.~Lin, X.~Zhao, and Z.~Shou.
\newblock Single shot temporal action detection.
\newblock In {\em ACM Multimedia}, 2017.

\bibitem{lin:eccv2014}
T.-Y. Lin, M.~Maire, S.~Belongie, J.~Hays, P.~Perona, D.~Ramanan,
  P.~Doll{\'a}r, and C.~Zitnick.
\newblock Microsoft {COCO}: Common objects in context.
\newblock In {\em ECCV}. 2014.

\bibitem{ma:cvpr2016}
S.~Ma, L.~Sigal, and S.~Sclaroff.
\newblock Learning activity progression in {LSTM}s for activity detection and
  early detection.
\newblock In {\em CVPR}, 2016.

\bibitem{ng:cvpr2015}
J.~Y.-H. Ng, M.~Hausknecht, S.~Vijayanarasimhan, O.~Vinyals, R.~Monga, and
  G.~Toderici.
\newblock Beyond short snippets: Deep networks for video classification.
\newblock In {\em CVPR}, 2015.

\bibitem{ni:cvpr2016}
B.~Ni, X.~Yang, and S.~Gao.
\newblock Progressively parsing interactional objects for fine grained action
  detection.
\newblock In {\em CVPR}, 2016.

\bibitem{oneata:2014}
D.~Oneata, J.~Verbeek, , and C.~Schmid.
\newblock The {LEAR} submission at thumos 2014.
\newblock \url{http://crcv.ucf.edu/THUMOS14/}, 2014.

\bibitem{ren:nips2015}
S.~Ren, K.~He, R.~Girshick, and J.~Sun.
\newblock Faster {R}-{CNN}: Towards real-time object detection with region
  proposal networks.
\newblock In {\em NIPS}. 2015.

\bibitem{richard:cvpr2016}
A.~Richard and J.~Gall.
\newblock Temporal action detection using a statistical language model.
\newblock In {\em CVPR}, 2016.

\bibitem{shou:cvpr2017}
Z.~Shou, J.~Chan, A.~Zareian, K.~Miyazawa, and S.-F. Chang.
\newblock Cdc: Convolutional-de-convolutional networks for precise temporal
  action localization in untrimmed videos.
\newblock In {\em CVPR}, 2017.

\bibitem{shou:cvpr2016}
Z.~Shou, D.~Wang, and S.-F. Chang.
\newblock Temporal action localization in untrimmed videos via multi-stage
  {CNN}s.
\newblock In {\em CVPR}, 2016.

\bibitem{simonyan:nips2014}
K.~Simonyan and A.~Zisserman.
\newblock Two-stream convolutional networks for action recognition in videos.
\newblock In {\em NIPS}. 2014.

\bibitem{singh:cvpr2016}
B.~Singh, T.~K. Marks, M.~Jones, O.~Tuzel, and M.~Shao.
\newblock A multi-stream bi-directional recurrent neural network for
  fine-grained action detection.
\newblock In {\em CVPR}, 2016.

\bibitem{singh:2016}
G.~Singh and F.~Cuzzolin.
\newblock Untrimmed video classification for activity detection: submission to
  {A}ctivity{N}et challenge.
\newblock In {\em {A}ctivity{N}et Large Scale Activity Recognition Challenge},
  2016.

\bibitem{singh:iccv2017}
G.~Singh, S.~Saha, M.~Sapienza, P.~Torr, and F.~Cuzzolin.
\newblock Online real-time multiple spatiotemporal action localisation and
  prediction.
\newblock In {\em ICCV}, 2017.

\bibitem{sun:acmmm2015}
C.~Sun, S.~Shetty, R.~Sukthankar, and R.~Nevatia.
\newblock Temporal localization of fine-grained actions in videos by domain
  transfer from web images.
\newblock In {\em ACM Multimedia}, 2015.

\bibitem{szegedy:cvpr2015}
C.~Szegedy, W.~Liu, Y.~Jia, P.~Sermanet, S.~Reed, D.~Anguelov, D.~Erhan,
  V.~Vanhoucke, and A.~Rabinovich.
\newblock Going deeper with convolutions.
\newblock In {\em CVPR}, 2015.

\bibitem{tran:iccv2015}
D.~Tran, L.~Bourdev, R.~Fergus, L.~Torresani, and M.~Paluri.
\newblock Learning spatiotemporal features with 3{D} convolutional networks.
\newblock In {\em ICCV}, 2015.

\bibitem{vijayanarasimhan:arxiv2017}
S.~Vijayanarasimhan, S.~Ricco, C.~Schmid, R.~Sukthankar, and K.~Fragkiadaki.
\newblock {S}f{M}-{N}et: Learning of structure and motion from video.
\newblock {\em arXiv preprint arXiv:1704.07804}, 2017.

\bibitem{vondrick:ijcv2013}
C.~Vondrick, D.~Patterson, and D.~Ramanan.
\newblock Efficiently scaling up crowdsourced video annotation.
\newblock {\em IJCV}, 101(1):184--204, Jan 2013.

\bibitem{wang:iccv2013}
H.~Wang and C.~Schmid.
\newblock Action recognition with improved trajectories.
\newblock In {\em ICCV}, 2013.

\bibitem{wang:2014}
L.~Wang, Y.~Qiao, and X.~Tang.
\newblock Action recognition and detection by combining motion and appearance
  features.
\newblock \url{http://crcv.ucf.edu/THUMOS14/}, 2014.

\bibitem{wang:cvpr2017}
L.~Wang, Y.~Xiong, D.~Lin, and L.~{Van Gool}.
\newblock Untrimmednets for weakly supervised action recognition and detection.
\newblock In {\em CVPR}, 2017.

\bibitem{wang:eccv2016}
L.~Wang, Y.~Xiong, Z.~Wang, Y.~Qiao, D.~Lin, X.~Tang, and L.~{Van Gool}.
\newblock Temporal segment networks: Towards good practices for deep action
  recognition.
\newblock In {\em ECCV}, 2016.

\bibitem{wang:2016}
R.~Wang and D.~Tao.
\newblock {UTS} at {A}ctivity{N}et 2016.
\newblock In {\em {A}ctivity{N}et Large Scale Activity Recognition Challenge},
  2016.

\bibitem{xu:iccv2017}
H.~Xu, A.~Das, and K.~Saenko.
\newblock R-{C}3{D}: Region convolutional 3{D} network for temporal activity
  detection.
\newblock In {\em ICCV}, 2017.

\bibitem{yeung:cvpr2016}
S.~Yeung, O.~Russakovsky, G.~Mori, and L.~Fei-Fei.
\newblock End-to-end learning of action detection from frame glimpses in
  videos.
\newblock In {\em CVPR}, 2016.

\bibitem{yu:iclr2016}
F.~Yu and V.~Koltun.
\newblock Multi-scale context aggregation by dilated convolutions.
\newblock In {\em ICLR}, 2016.

\bibitem{yuan:cvpr2016}
J.~Yuan, B.~Ni, X.~Yang, and A.~A. Kassim.
\newblock Temporal action localization with pyramid of score distribution
  features.
\newblock In {\em CVPR}, 2016.

\bibitem{yuan:cvpr2017}
Z.~Yuan, J.~C. Stroud, T.~Lu, and J.~Deng.
\newblock Temporal action localization by structured maximal sums.
\newblock In {\em CVPR}, 2017.

\bibitem{zhao:iccv2017}
Y.~Zhao, Y.~Xiong, L.~Wang, Z.~Wu, D.~Lin, and X.~Tang.
\newblock Temporal action detection with structured segment networks.
\newblock In {\em ICCV}, 2017.

\end{thebibliography}
}

\clearpage

\appendix

\section{Supplementary Material}

\subsection{Training Strategy}

As in Faster R-CNN~\cite{ren:nips2015}, the training of proposal generation and
action classification share a same form of multi-task loss, targeting both
classification and regression:
\begin{equation}
 \mathcal{L}=\sum_i \mathcal{L}_{cls}(p_i,p^*_i)+\lambda\sum_i [p^*_i\ge1] \mathcal{L}_{reg}(t_i,t^*_i).
\end{equation}
$i$ is the index of an anchor or proposal in a mini-batch. For classification,
$p$ is the predicted probability of the proposal or actions, $p^*$ is the
ground-truth label, and $\mathcal{L}_{cls}$ is the cross-entropy loss. Note
that $p^*\in\{0,1\}$ for proposal generation, and $p^*\in\{0,\ldots,C\}$ for
action classification, where $C$ is the number of action classes of interest
and 0 accounts for the background action class. For regression, $t$ is the
predicted offset relative to an anchor or proposal, $t^*$ is the ground-truth
offset, and $\mathcal{L}_{reg}$ is the smooth L1 loss defined in
\cite{girshick:iccv2015}. We parameterize the offsets $t=(t_c,t_l)$ and
$t^*=(t^*_c,t^*_l)$ by:
\begin{equation}
 \begin{aligned}
  &t_c=10\cdot(c-c_a)/c_i,&&t_l=5\cdot\log(l/l_a), \\
  &t^*_c=10\cdot(c^*-c_a)/c_i,&&t^*_l=5\cdot\log(l^*/l_a),
 \end{aligned}
\end{equation}
where $c$ and $l$ denote the segment's center coordinate and its length. $c$
and $c^*$ account for the predicted and ground-truth segments, while $c_a$
accounts for the anchor and proposal segments, for proposal generation and
action classification, respectively (similarly for $l$). The indicator function
$[\cdot]$ is used to exclude the background anchors and proposals when the
regression loss is computed. In all experiments, we set $\lambda=1$ for both
proposal generation and action classification, and jointly train both stages by
weighing both losses equally.

For proposal generation, an anchor is assigned a positive label if it overlaps
with a ground-truth segment with temporal Intersection-over-Union (tIoU) higher
than 0.7. A negative label is assigned if the tIoU overlap is lower than 0.3
with all ground-truth segments. We also force each ground-truth segment to have
at least one matched positive anchor. For action classification, a proposal is
assigned the action label of its most overlapped ground-truth segment, if the
ground-truth segment has tIoU overlap over 0.5. Otherwise a background label
(i.e. 0) is assigned.

Each mini-batch contains examples sampled from a single video. For proposal
generation, we set the mini-batch size to 256 and the fraction of positives to
0.5. For action classification, we set the mini-batch size to 64 and the
fraction of foreground actions to 0.25. We use the Adam optimizer with a
learning rate of 0.0001.

\subsection{Additional Qualitative Results}

\begin{figure*}[t]
 \centering
 \begin{minipage}{0.058\textwidth} \centering \includegraphics[width=1.00\textwidth]{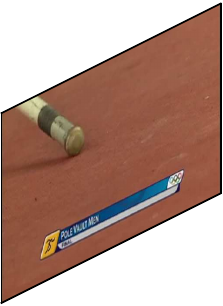} \end{minipage}
 \begin{minipage}{0.058\textwidth} \centering \includegraphics[width=1.00\textwidth]{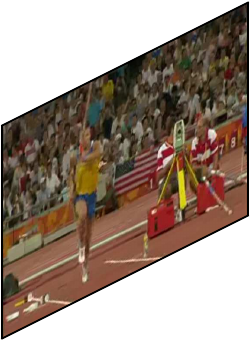} \end{minipage}
 \begin{minipage}{0.058\textwidth} \centering \includegraphics[width=1.00\textwidth]{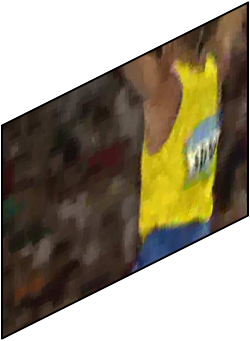} \end{minipage}
 \begin{minipage}{0.058\textwidth} \centering \includegraphics[width=1.00\textwidth]{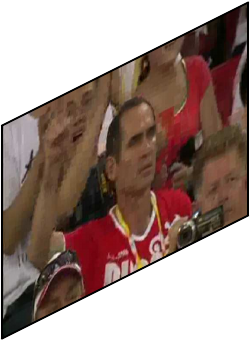} \end{minipage}
 \begin{minipage}{0.058\textwidth} \centering \includegraphics[width=1.00\textwidth]{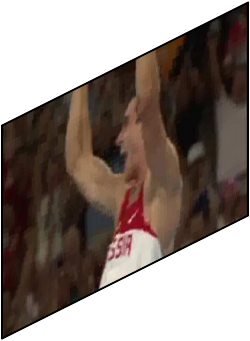} \end{minipage}
 \begin{minipage}{0.058\textwidth} \centering \includegraphics[width=1.00\textwidth]{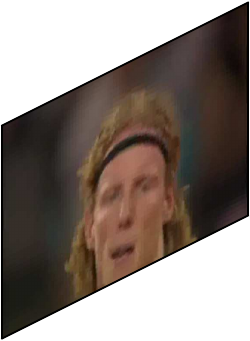} \end{minipage}
 \begin{minipage}{0.058\textwidth} \centering \includegraphics[width=1.00\textwidth]{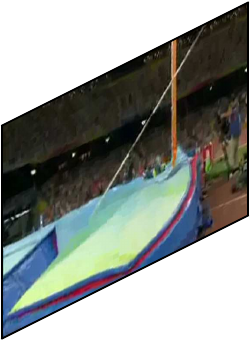} \end{minipage}
 \begin{minipage}{0.058\textwidth} \centering \includegraphics[width=1.00\textwidth]{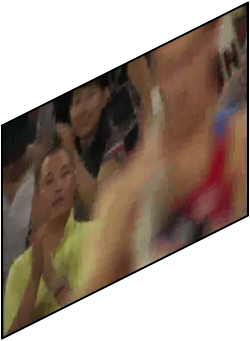} \end{minipage}
 \begin{minipage}{0.058\textwidth} \centering \includegraphics[width=1.00\textwidth]{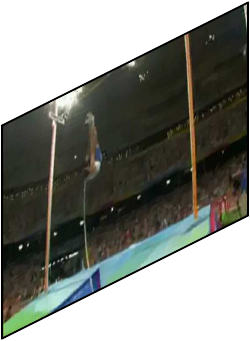} \end{minipage}
 \begin{minipage}{0.058\textwidth} \centering \includegraphics[width=1.00\textwidth]{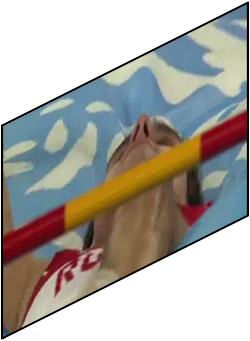} \end{minipage}
 \begin{minipage}{0.058\textwidth} \centering \includegraphics[width=1.00\textwidth]{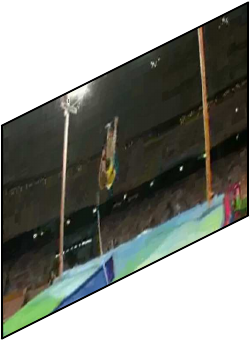} \end{minipage}
 \begin{minipage}{0.058\textwidth} \centering \includegraphics[width=1.00\textwidth]{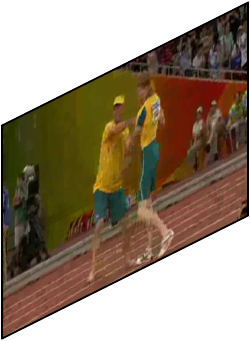} \end{minipage}
 \begin{minipage}{0.058\textwidth} \centering \includegraphics[width=1.00\textwidth]{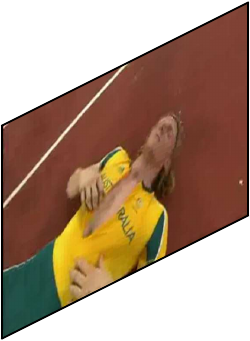} \end{minipage}
 \begin{minipage}{0.058\textwidth} \centering \includegraphics[width=1.00\textwidth]{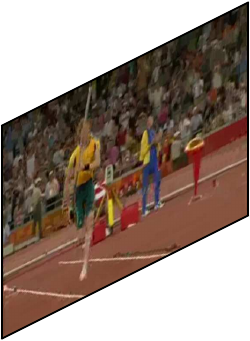} \end{minipage}
 \begin{minipage}{0.058\textwidth} \centering \includegraphics[width=1.00\textwidth]{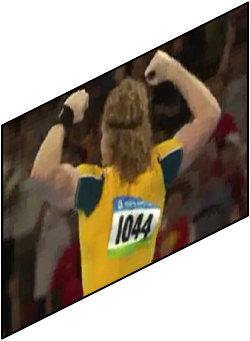} \end{minipage}
 \begin{minipage}{0.058\textwidth} \centering \includegraphics[width=1.00\textwidth]{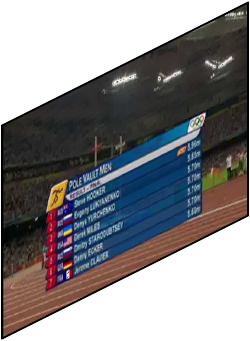} \end{minipage}
 \\ \vspace{2mm}
 \includegraphics[width=1.00\textwidth]{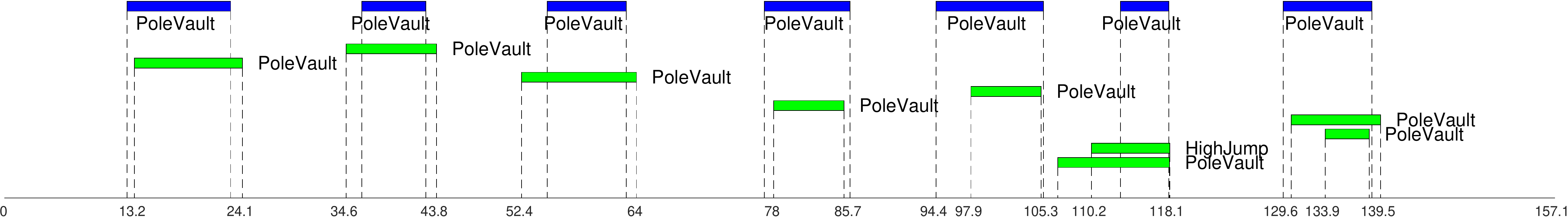}
 \\
 (a)
 \\ \vspace{2mm}
 \begin{minipage}{0.058\textwidth} \centering \includegraphics[width=1.00\textwidth]{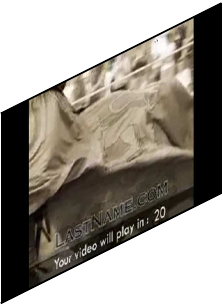} \end{minipage}
 \begin{minipage}{0.058\textwidth} \centering \includegraphics[width=1.00\textwidth]{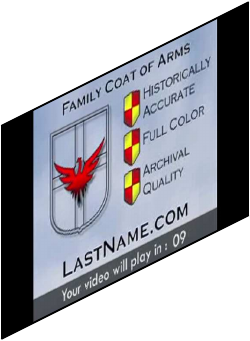} \end{minipage}
 \begin{minipage}{0.058\textwidth} \centering \includegraphics[width=1.00\textwidth]{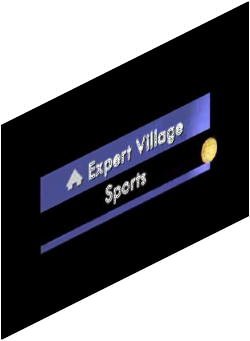} \end{minipage}
 \begin{minipage}{0.058\textwidth} \centering \includegraphics[width=1.00\textwidth]{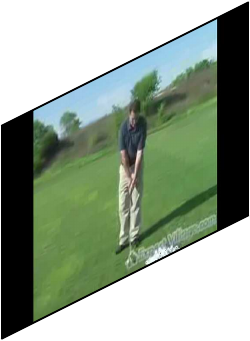} \end{minipage}
 \begin{minipage}{0.058\textwidth} \centering \includegraphics[width=1.00\textwidth]{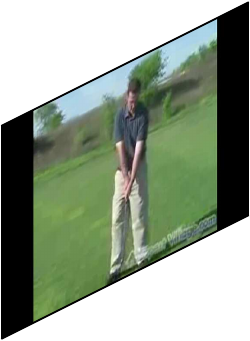} \end{minipage}
 \begin{minipage}{0.058\textwidth} \centering \includegraphics[width=1.00\textwidth]{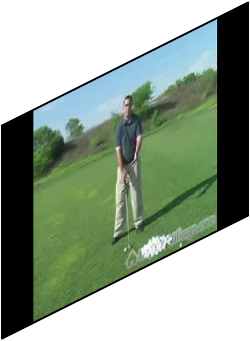} \end{minipage}
 \begin{minipage}{0.058\textwidth} \centering \includegraphics[width=1.00\textwidth]{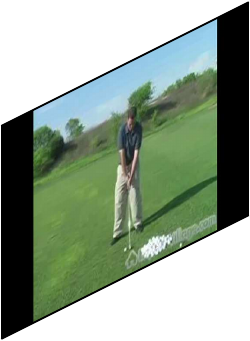} \end{minipage}
 \begin{minipage}{0.058\textwidth} \centering \includegraphics[width=1.00\textwidth]{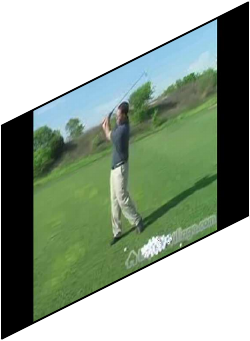} \end{minipage}
 \begin{minipage}{0.058\textwidth} \centering \includegraphics[width=1.00\textwidth]{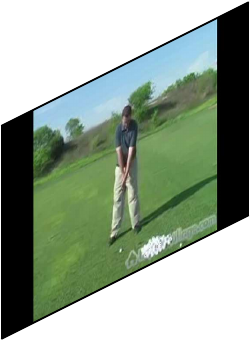} \end{minipage}
 \begin{minipage}{0.058\textwidth} \centering \includegraphics[width=1.00\textwidth]{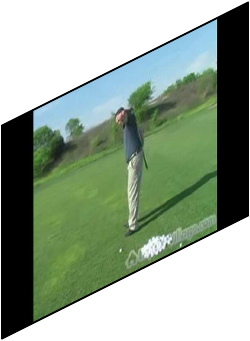} \end{minipage}
 \begin{minipage}{0.058\textwidth} \centering \includegraphics[width=1.00\textwidth]{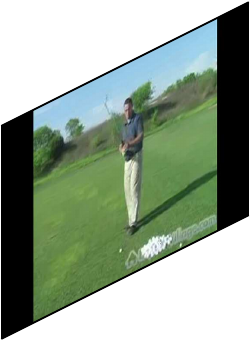} \end{minipage}
 \begin{minipage}{0.058\textwidth} \centering \includegraphics[width=1.00\textwidth]{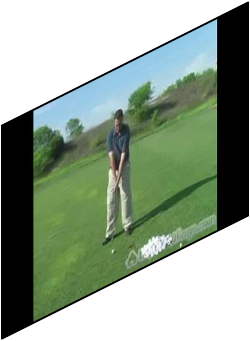} \end{minipage}
 \begin{minipage}{0.058\textwidth} \centering \includegraphics[width=1.00\textwidth]{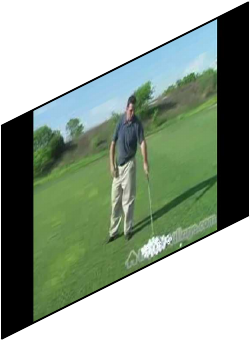} \end{minipage}
 \begin{minipage}{0.058\textwidth} \centering \includegraphics[width=1.00\textwidth]{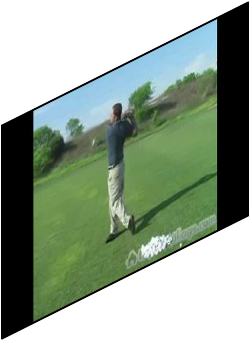} \end{minipage}
 \begin{minipage}{0.058\textwidth} \centering \includegraphics[width=1.00\textwidth]{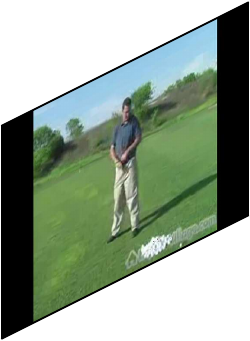} \end{minipage}
 \begin{minipage}{0.058\textwidth} \centering \includegraphics[width=1.00\textwidth]{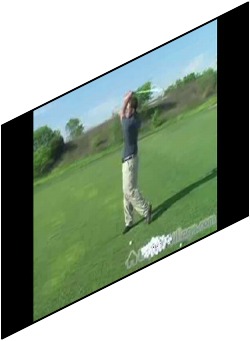} \end{minipage}
 \\ \vspace{2mm}
 \includegraphics[width=1.00\textwidth]{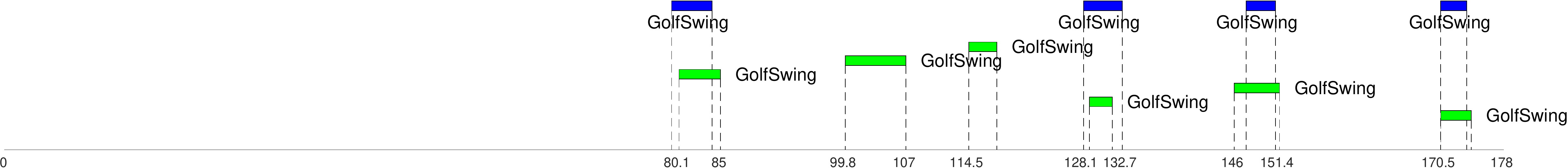}
 \\
 (b)
 \\ \vspace{2mm}
 \begin{minipage}{0.058\textwidth} \centering \includegraphics[width=1.00\textwidth]{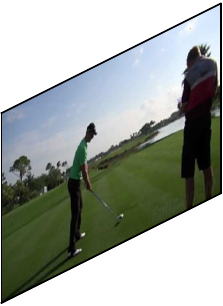} \end{minipage}
 \begin{minipage}{0.058\textwidth} \centering \includegraphics[width=1.00\textwidth]{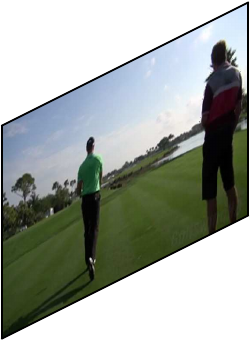} \end{minipage}
 \begin{minipage}{0.058\textwidth} \centering \includegraphics[width=1.00\textwidth]{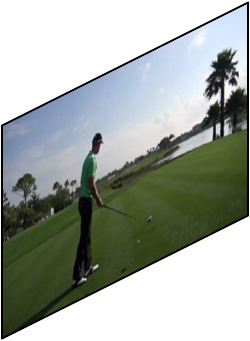} \end{minipage}
 \begin{minipage}{0.058\textwidth} \centering \includegraphics[width=1.00\textwidth]{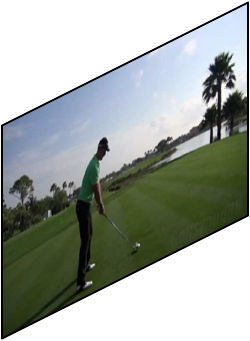} \end{minipage}
 \begin{minipage}{0.058\textwidth} \centering \includegraphics[width=1.00\textwidth]{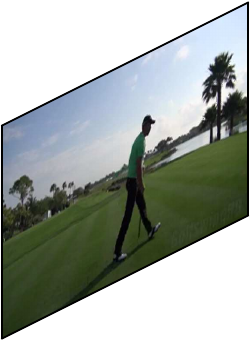} \end{minipage}
 \begin{minipage}{0.058\textwidth} \centering \includegraphics[width=1.00\textwidth]{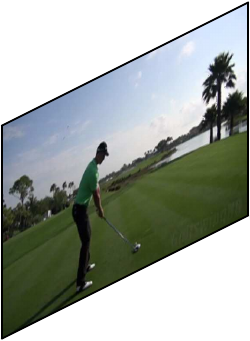} \end{minipage}
 \begin{minipage}{0.058\textwidth} \centering \includegraphics[width=1.00\textwidth]{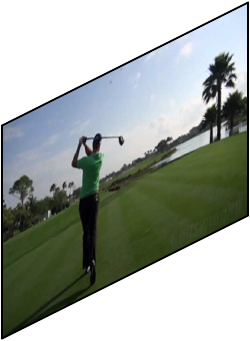} \end{minipage}
 \begin{minipage}{0.058\textwidth} \centering \includegraphics[width=1.00\textwidth]{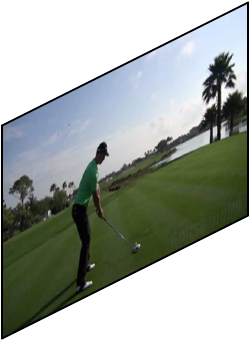} \end{minipage}
 \begin{minipage}{0.058\textwidth} \centering \includegraphics[width=1.00\textwidth]{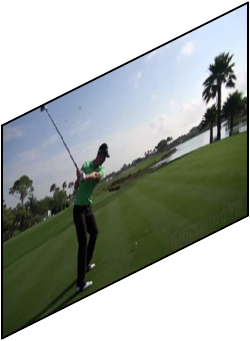} \end{minipage}
 \begin{minipage}{0.058\textwidth} \centering \includegraphics[width=1.00\textwidth]{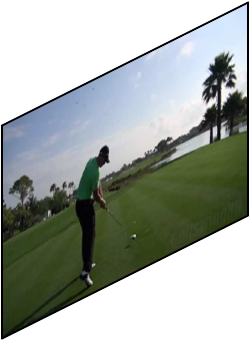} \end{minipage}
 \begin{minipage}{0.058\textwidth} \centering \includegraphics[width=1.00\textwidth]{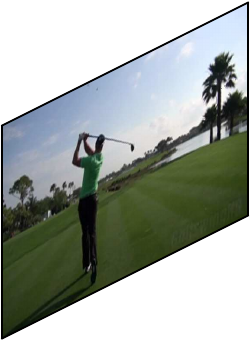} \end{minipage}
 \begin{minipage}{0.058\textwidth} \centering \includegraphics[width=1.00\textwidth]{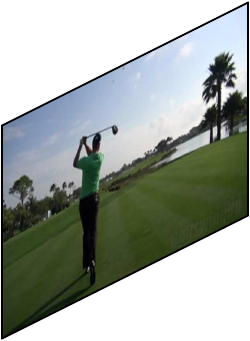} \end{minipage}
 \begin{minipage}{0.058\textwidth} \centering \includegraphics[width=1.00\textwidth]{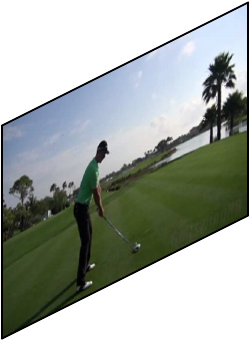} \end{minipage}
 \begin{minipage}{0.058\textwidth} \centering \includegraphics[width=1.00\textwidth]{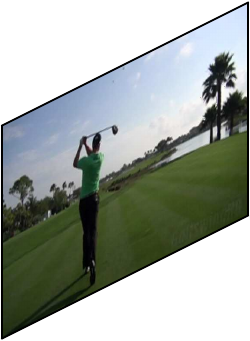} \end{minipage}
 \begin{minipage}{0.058\textwidth} \centering \includegraphics[width=1.00\textwidth]{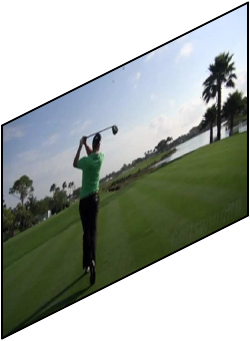} \end{minipage}
 \begin{minipage}{0.058\textwidth} \centering \includegraphics[width=1.00\textwidth]{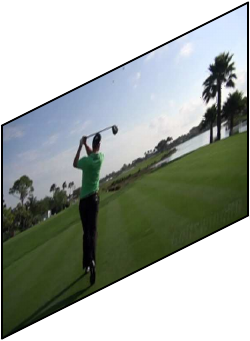} \end{minipage}
 \\ \vspace{2mm}
 \includegraphics[width=1.00\textwidth]{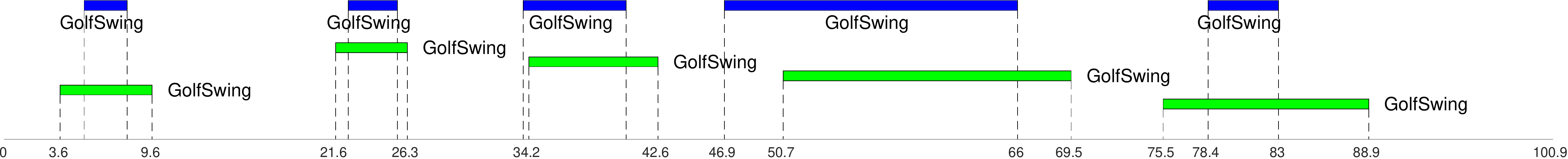}
 \\
 (c)
 \\ \vspace{2mm}
 \begin{minipage}{0.058\textwidth} \centering \includegraphics[width=1.00\textwidth]{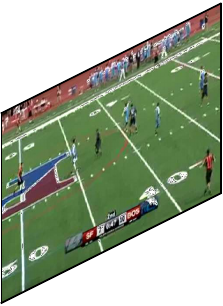} \end{minipage}
 \begin{minipage}{0.058\textwidth} \centering \includegraphics[width=1.00\textwidth]{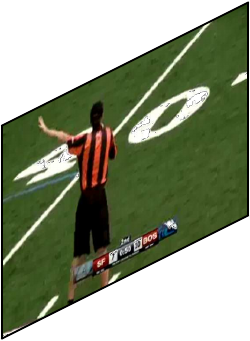} \end{minipage}
 \begin{minipage}{0.058\textwidth} \centering \includegraphics[width=1.00\textwidth]{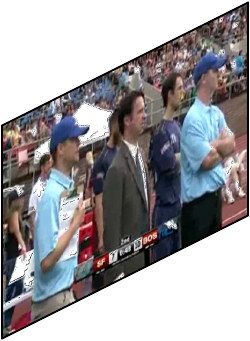} \end{minipage}
 \begin{minipage}{0.058\textwidth} \centering \includegraphics[width=1.00\textwidth]{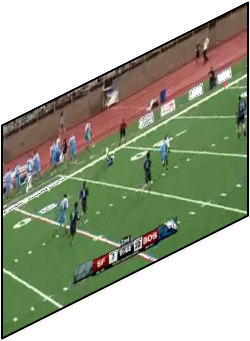} \end{minipage}
 \begin{minipage}{0.058\textwidth} \centering \includegraphics[width=1.00\textwidth]{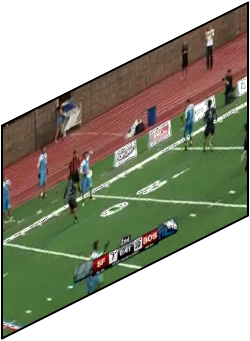} \end{minipage}
 \begin{minipage}{0.058\textwidth} \centering \includegraphics[width=1.00\textwidth]{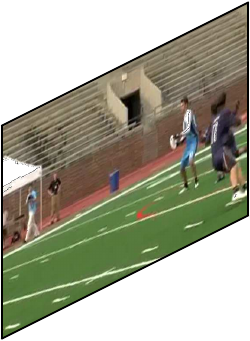} \end{minipage}
 \begin{minipage}{0.058\textwidth} \centering \includegraphics[width=1.00\textwidth]{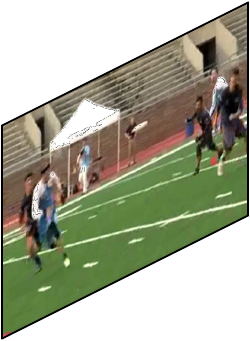} \end{minipage}
 \begin{minipage}{0.058\textwidth} \centering \includegraphics[width=1.00\textwidth]{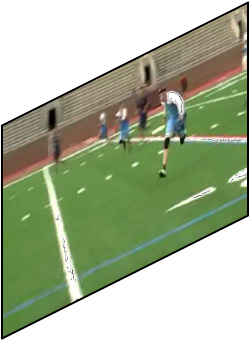} \end{minipage}
 \begin{minipage}{0.058\textwidth} \centering \includegraphics[width=1.00\textwidth]{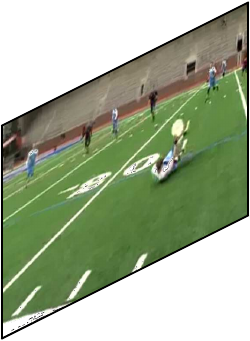} \end{minipage}
 \begin{minipage}{0.058\textwidth} \centering \includegraphics[width=1.00\textwidth]{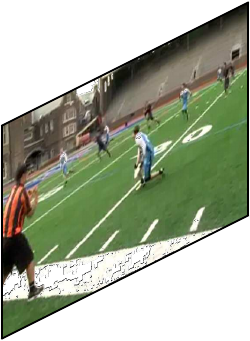} \end{minipage}
 \begin{minipage}{0.058\textwidth} \centering \includegraphics[width=1.00\textwidth]{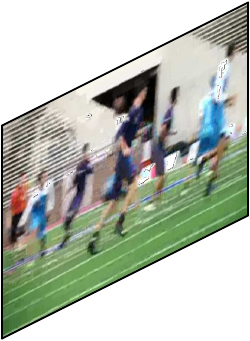} \end{minipage}
 \begin{minipage}{0.058\textwidth} \centering \includegraphics[width=1.00\textwidth]{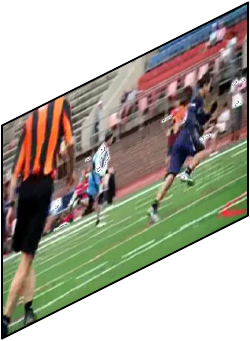} \end{minipage}
 \begin{minipage}{0.058\textwidth} \centering \includegraphics[width=1.00\textwidth]{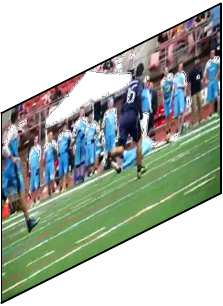} \end{minipage}
 \begin{minipage}{0.058\textwidth} \centering \includegraphics[width=1.00\textwidth]{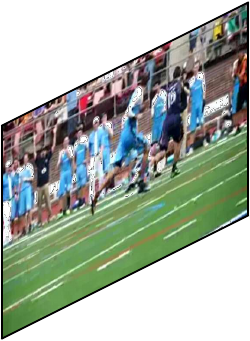} \end{minipage}
 \begin{minipage}{0.058\textwidth} \centering \includegraphics[width=1.00\textwidth]{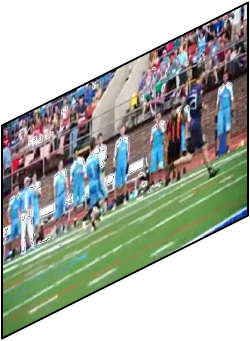} \end{minipage}
 \begin{minipage}{0.058\textwidth} \centering \includegraphics[width=1.00\textwidth]{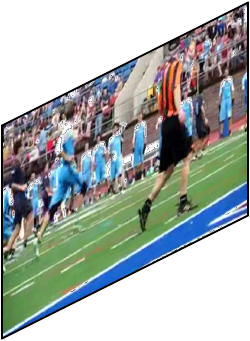} \end{minipage}
 \\ \vspace{2mm}
 \includegraphics[width=1.00\textwidth]{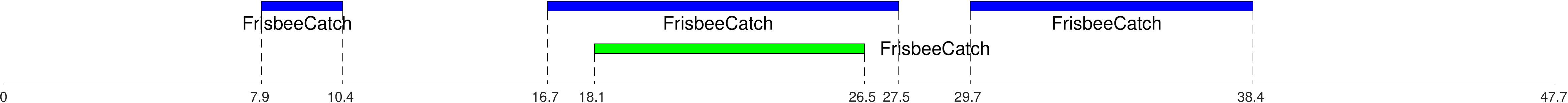}
 \\
 (d)
 \\ \vspace{2mm}
 \begin{minipage}{0.058\textwidth} \centering \includegraphics[width=1.00\textwidth]{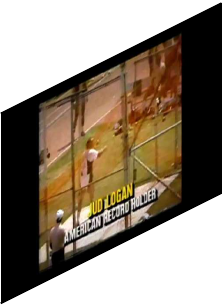} \end{minipage}
 \begin{minipage}{0.058\textwidth} \centering \includegraphics[width=1.00\textwidth]{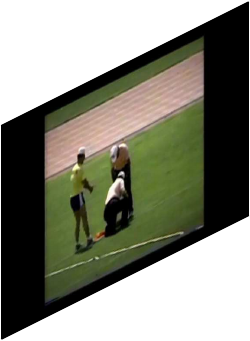} \end{minipage}
 \begin{minipage}{0.058\textwidth} \centering \includegraphics[width=1.00\textwidth]{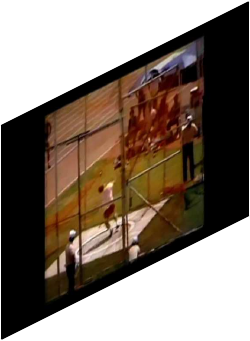} \end{minipage}
 \begin{minipage}{0.058\textwidth} \centering \includegraphics[width=1.00\textwidth]{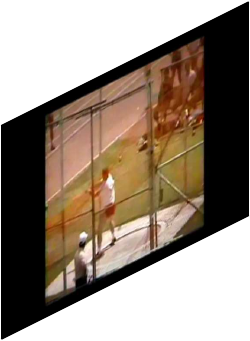} \end{minipage}
 \begin{minipage}{0.058\textwidth} \centering \includegraphics[width=1.00\textwidth]{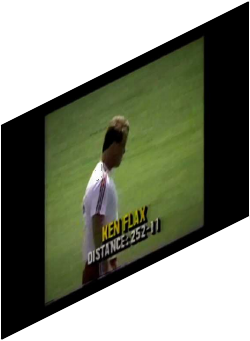} \end{minipage}
 \begin{minipage}{0.058\textwidth} \centering \includegraphics[width=1.00\textwidth]{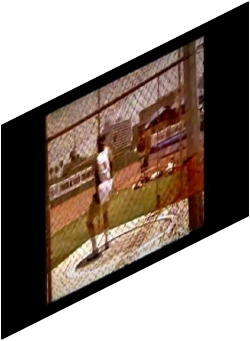} \end{minipage}
 \begin{minipage}{0.058\textwidth} \centering \includegraphics[width=1.00\textwidth]{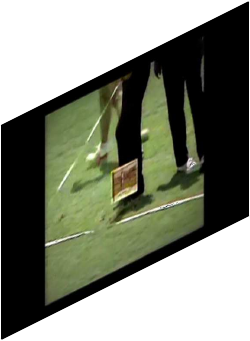} \end{minipage}
 \begin{minipage}{0.058\textwidth} \centering \includegraphics[width=1.00\textwidth]{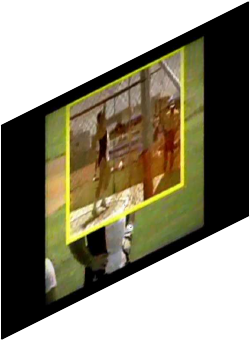} \end{minipage}
 \begin{minipage}{0.058\textwidth} \centering \includegraphics[width=1.00\textwidth]{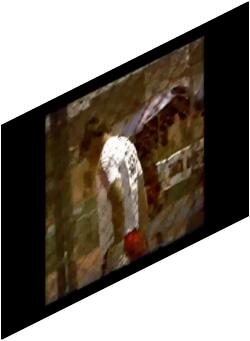} \end{minipage}
 \begin{minipage}{0.058\textwidth} \centering \includegraphics[width=1.00\textwidth]{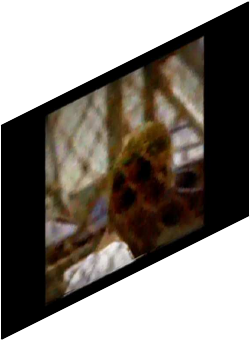} \end{minipage}
 \begin{minipage}{0.058\textwidth} \centering \includegraphics[width=1.00\textwidth]{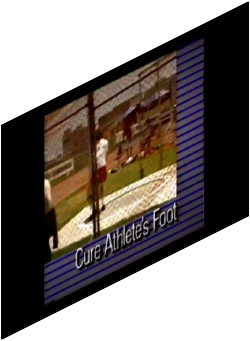} \end{minipage}
 \begin{minipage}{0.058\textwidth} \centering \includegraphics[width=1.00\textwidth]{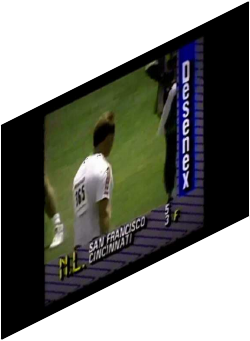} \end{minipage}
 \begin{minipage}{0.058\textwidth} \centering \includegraphics[width=1.00\textwidth]{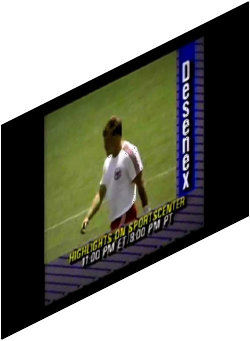} \end{minipage}
 \begin{minipage}{0.058\textwidth} \centering \includegraphics[width=1.00\textwidth]{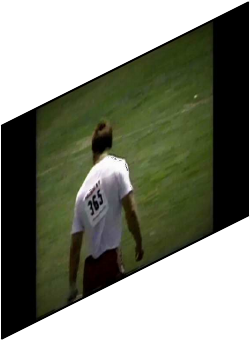} \end{minipage}
 \begin{minipage}{0.058\textwidth} \centering \includegraphics[width=1.00\textwidth]{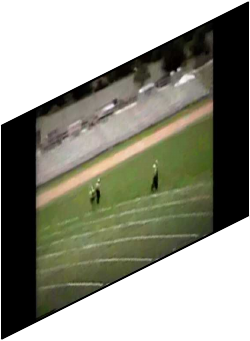} \end{minipage}
 \begin{minipage}{0.058\textwidth} \centering \includegraphics[width=1.00\textwidth]{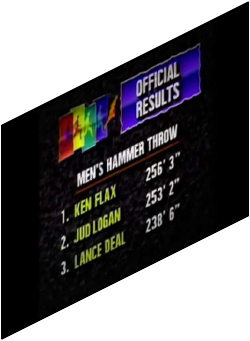} \end{minipage}
 \\ \vspace{2mm}
 \includegraphics[width=1.00\textwidth]{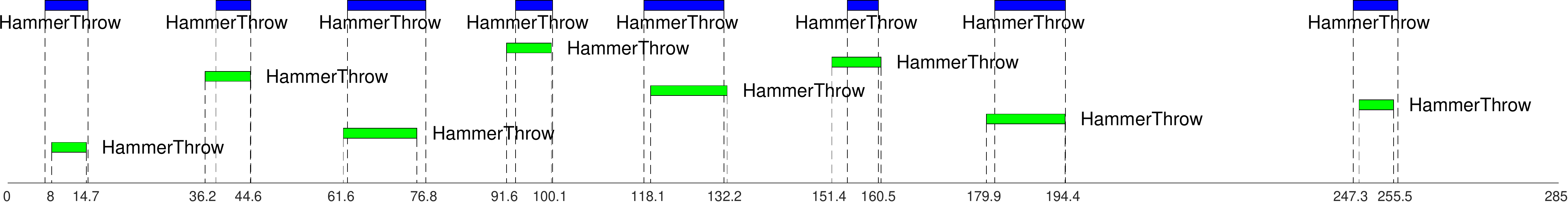}
 \\
 (e)
 \caption{\small Additional qualitative examples of the top localized actions
on THUMOS'14. Each consists of a sequence of frames sampled from a full test
video, the ground-truth (blue) and predicted (green) action segments and class
labels, and a temporal axis showing the time in seconds.}
 \label{fig:additional-1}
\end{figure*}

\begin{figure*}[t]
 \centering
 \begin{minipage}{0.058\textwidth} \centering \includegraphics[width=1.00\textwidth]{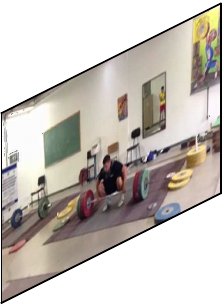} \end{minipage}
 \begin{minipage}{0.058\textwidth} \centering \includegraphics[width=1.00\textwidth]{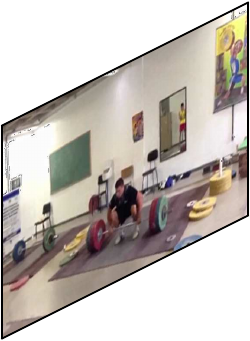} \end{minipage}
 \begin{minipage}{0.058\textwidth} \centering \includegraphics[width=1.00\textwidth]{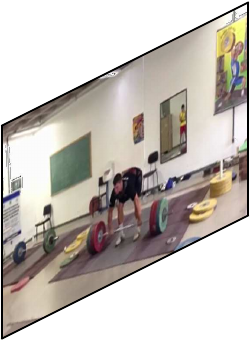} \end{minipage}
 \begin{minipage}{0.058\textwidth} \centering \includegraphics[width=1.00\textwidth]{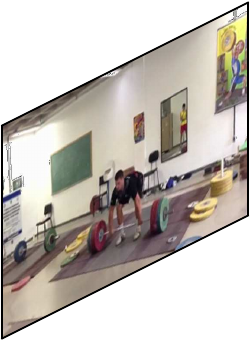} \end{minipage}
 \begin{minipage}{0.058\textwidth} \centering \includegraphics[width=1.00\textwidth]{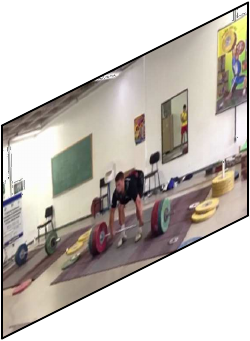} \end{minipage}
 \begin{minipage}{0.058\textwidth} \centering \includegraphics[width=1.00\textwidth]{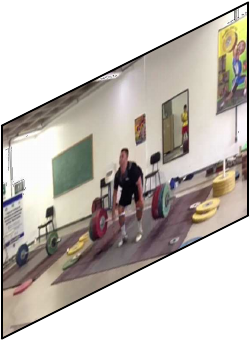} \end{minipage}
 \begin{minipage}{0.058\textwidth} \centering \includegraphics[width=1.00\textwidth]{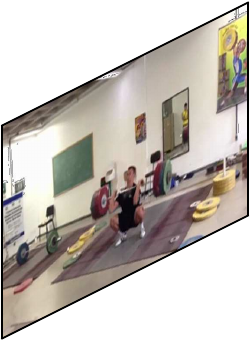} \end{minipage}
 \begin{minipage}{0.058\textwidth} \centering \includegraphics[width=1.00\textwidth]{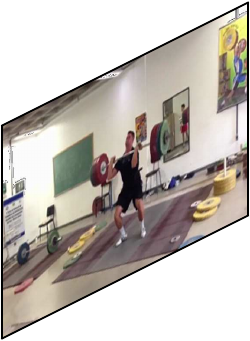} \end{minipage}
 \begin{minipage}{0.058\textwidth} \centering \includegraphics[width=1.00\textwidth]{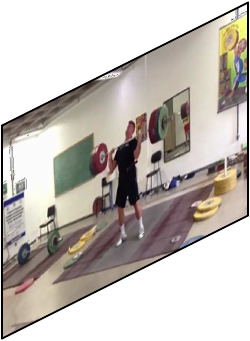} \end{minipage}
 \begin{minipage}{0.058\textwidth} \centering \includegraphics[width=1.00\textwidth]{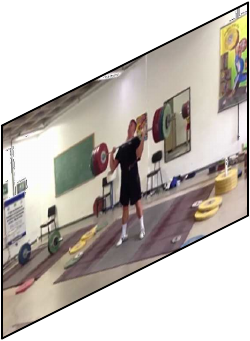} \end{minipage}
 \begin{minipage}{0.058\textwidth} \centering \includegraphics[width=1.00\textwidth]{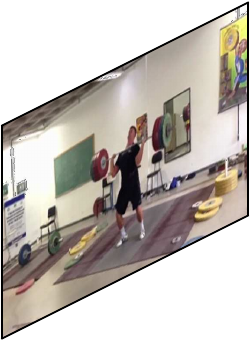} \end{minipage}
 \begin{minipage}{0.058\textwidth} \centering \includegraphics[width=1.00\textwidth]{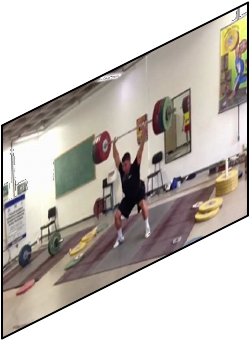} \end{minipage}
 \begin{minipage}{0.058\textwidth} \centering \includegraphics[width=1.00\textwidth]{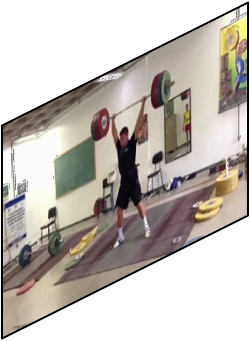} \end{minipage}
 \begin{minipage}{0.058\textwidth} \centering \includegraphics[width=1.00\textwidth]{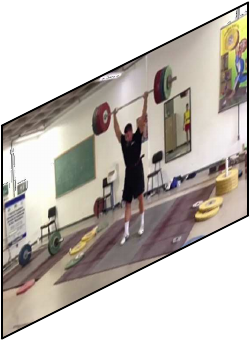} \end{minipage}
 \begin{minipage}{0.058\textwidth} \centering \includegraphics[width=1.00\textwidth]{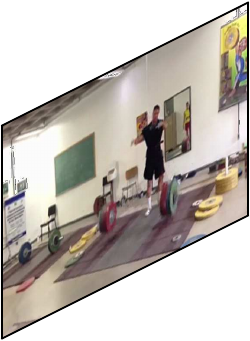} \end{minipage}
 \begin{minipage}{0.058\textwidth} \centering \includegraphics[width=1.00\textwidth]{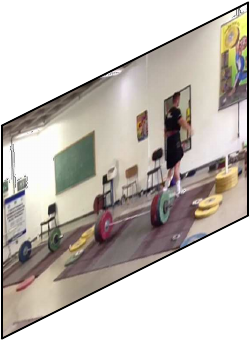} \end{minipage}
 \\ \vspace{2mm}
 \includegraphics[width=1.00\textwidth]{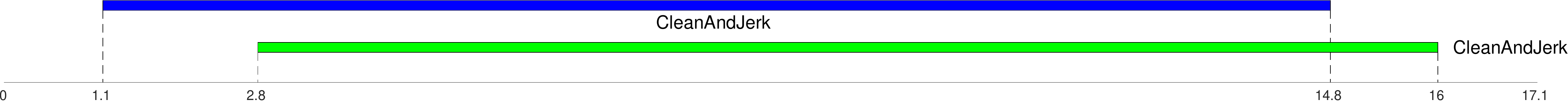}
 \\
 (a)
 \\ \vspace{2mm}
 \begin{minipage}{0.058\textwidth} \centering \includegraphics[width=1.00\textwidth]{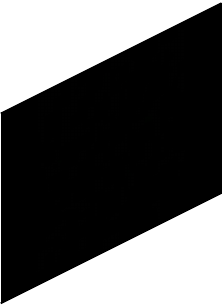} \end{minipage}
 \begin{minipage}{0.058\textwidth} \centering \includegraphics[width=1.00\textwidth]{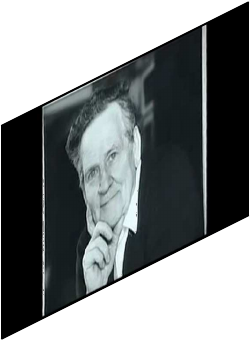} \end{minipage}
 \begin{minipage}{0.058\textwidth} \centering \includegraphics[width=1.00\textwidth]{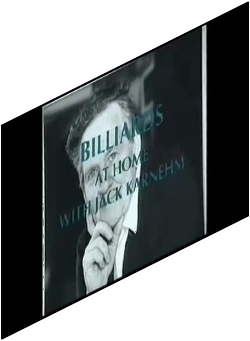} \end{minipage}
 \begin{minipage}{0.058\textwidth} \centering \includegraphics[width=1.00\textwidth]{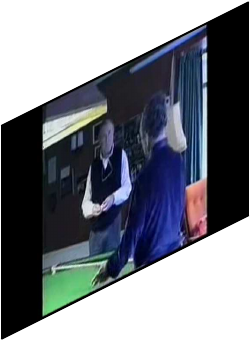} \end{minipage}
 \begin{minipage}{0.058\textwidth} \centering \includegraphics[width=1.00\textwidth]{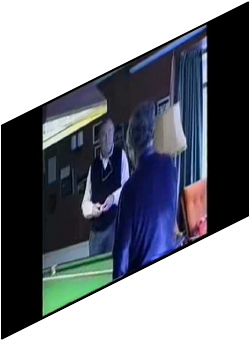} \end{minipage}
 \begin{minipage}{0.058\textwidth} \centering \includegraphics[width=1.00\textwidth]{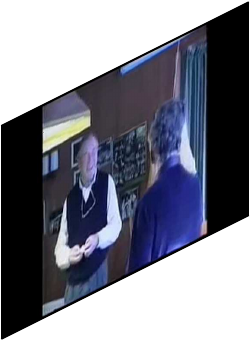} \end{minipage}
 \begin{minipage}{0.058\textwidth} \centering \includegraphics[width=1.00\textwidth]{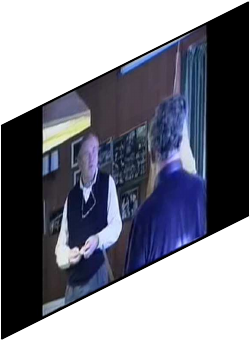} \end{minipage}
 \begin{minipage}{0.058\textwidth} \centering \includegraphics[width=1.00\textwidth]{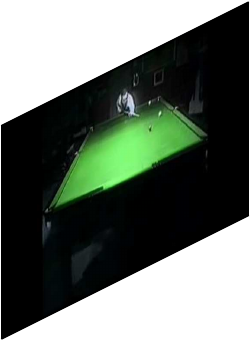} \end{minipage}
 \begin{minipage}{0.058\textwidth} \centering \includegraphics[width=1.00\textwidth]{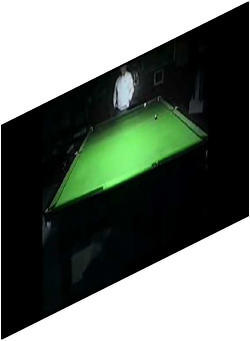} \end{minipage}
 \begin{minipage}{0.058\textwidth} \centering \includegraphics[width=1.00\textwidth]{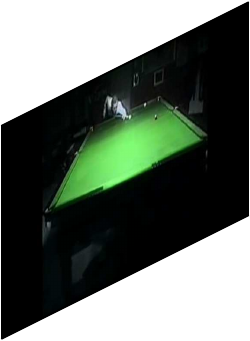} \end{minipage}
 \begin{minipage}{0.058\textwidth} \centering \includegraphics[width=1.00\textwidth]{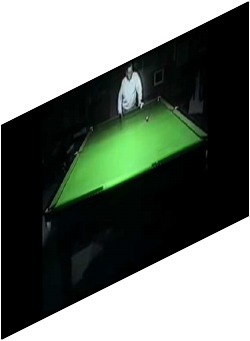} \end{minipage}
 \begin{minipage}{0.058\textwidth} \centering \includegraphics[width=1.00\textwidth]{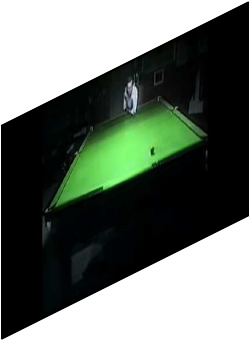} \end{minipage}
 \begin{minipage}{0.058\textwidth} \centering \includegraphics[width=1.00\textwidth]{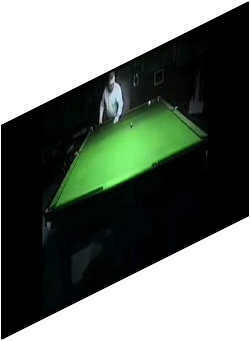} \end{minipage}
 \begin{minipage}{0.058\textwidth} \centering \includegraphics[width=1.00\textwidth]{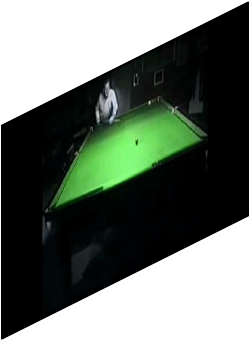} \end{minipage}
 \begin{minipage}{0.058\textwidth} \centering \includegraphics[width=1.00\textwidth]{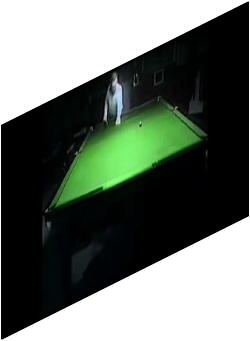} \end{minipage}
 \begin{minipage}{0.058\textwidth} \centering \includegraphics[width=1.00\textwidth]{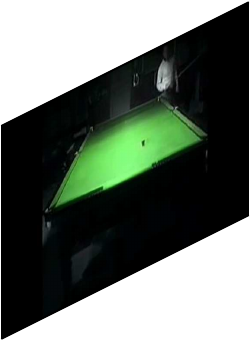} \end{minipage}
 \\ \vspace{2mm}
 \includegraphics[width=1.00\textwidth]{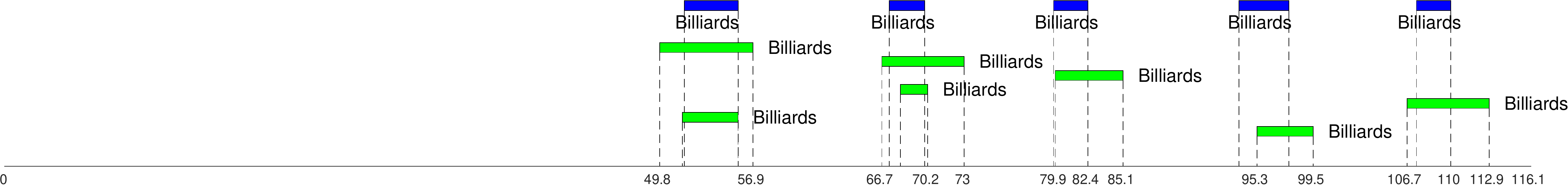}
 \\
 (b)
 \\ \vspace{2mm}
 \begin{minipage}{0.058\textwidth} \centering \includegraphics[width=1.00\textwidth]{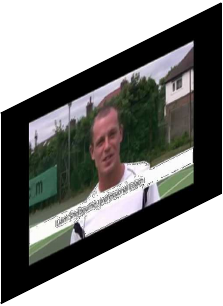} \end{minipage}
 \begin{minipage}{0.058\textwidth} \centering \includegraphics[width=1.00\textwidth]{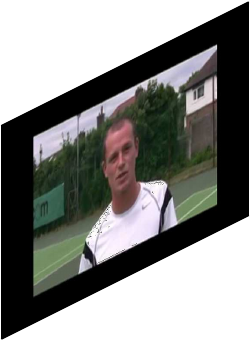} \end{minipage}
 \begin{minipage}{0.058\textwidth} \centering \includegraphics[width=1.00\textwidth]{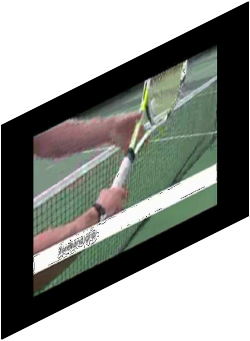} \end{minipage}
 \begin{minipage}{0.058\textwidth} \centering \includegraphics[width=1.00\textwidth]{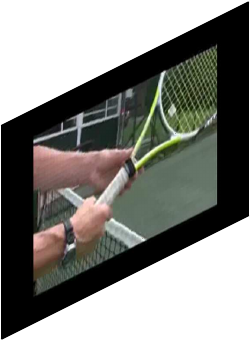} \end{minipage}
 \begin{minipage}{0.058\textwidth} \centering \includegraphics[width=1.00\textwidth]{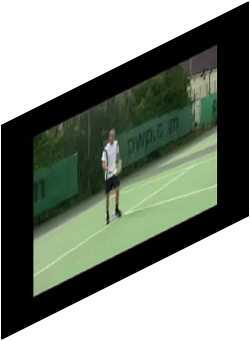} \end{minipage}
 \begin{minipage}{0.058\textwidth} \centering \includegraphics[width=1.00\textwidth]{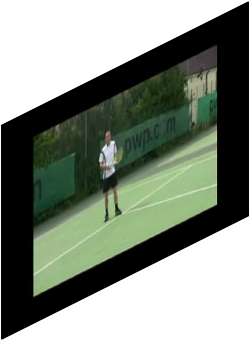} \end{minipage}
 \begin{minipage}{0.058\textwidth} \centering \includegraphics[width=1.00\textwidth]{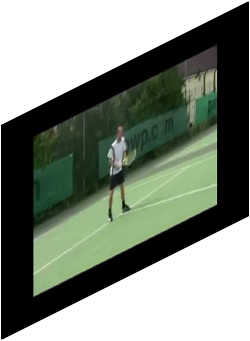} \end{minipage}
 \begin{minipage}{0.058\textwidth} \centering \includegraphics[width=1.00\textwidth]{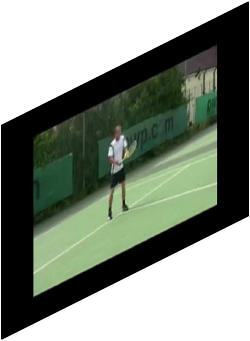} \end{minipage}
 \begin{minipage}{0.058\textwidth} \centering \includegraphics[width=1.00\textwidth]{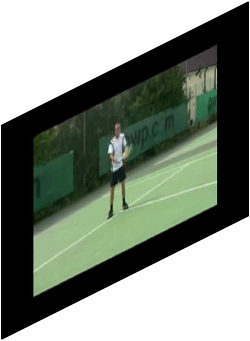} \end{minipage}
 \begin{minipage}{0.058\textwidth} \centering \includegraphics[width=1.00\textwidth]{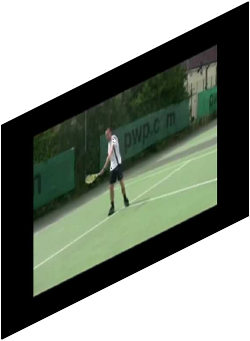} \end{minipage}
 \begin{minipage}{0.058\textwidth} \centering \includegraphics[width=1.00\textwidth]{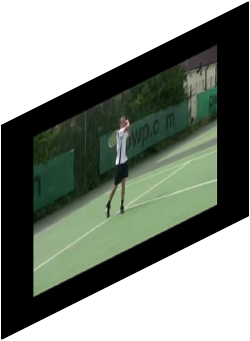} \end{minipage}
 \begin{minipage}{0.058\textwidth} \centering \includegraphics[width=1.00\textwidth]{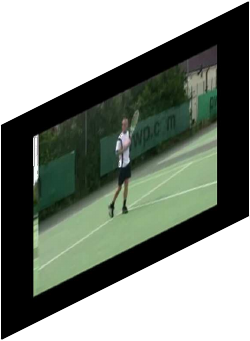} \end{minipage}
 \begin{minipage}{0.058\textwidth} \centering \includegraphics[width=1.00\textwidth]{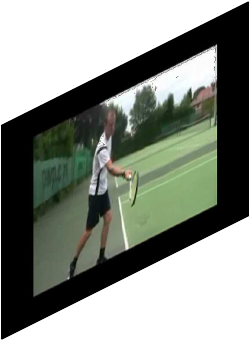} \end{minipage}
 \begin{minipage}{0.058\textwidth} \centering \includegraphics[width=1.00\textwidth]{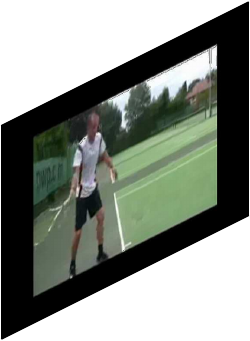} \end{minipage}
 \begin{minipage}{0.058\textwidth} \centering \includegraphics[width=1.00\textwidth]{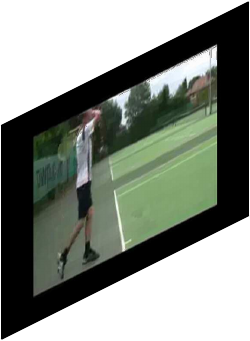} \end{minipage}
 \begin{minipage}{0.058\textwidth} \centering \includegraphics[width=1.00\textwidth]{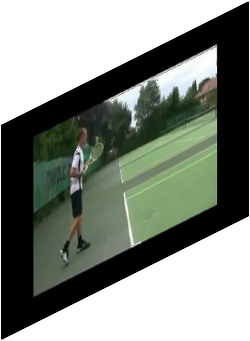} \end{minipage}
 \\ \vspace{2mm}
 \includegraphics[width=1.00\textwidth]{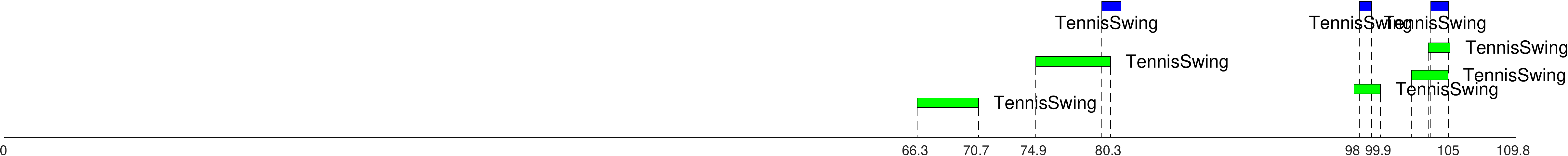}
 \\
 (c)
 \\ \vspace{2mm}
 \begin{minipage}{0.058\textwidth} \centering \includegraphics[width=1.00\textwidth]{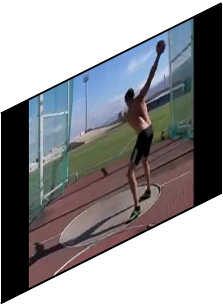} \end{minipage}
 \begin{minipage}{0.058\textwidth} \centering \includegraphics[width=1.00\textwidth]{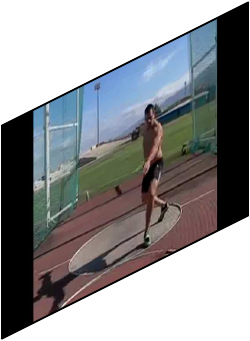} \end{minipage}
 \begin{minipage}{0.058\textwidth} \centering \includegraphics[width=1.00\textwidth]{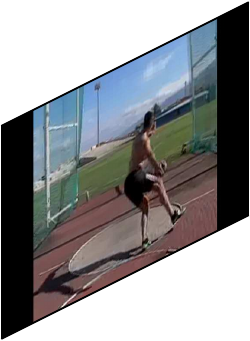} \end{minipage}
 \begin{minipage}{0.058\textwidth} \centering \includegraphics[width=1.00\textwidth]{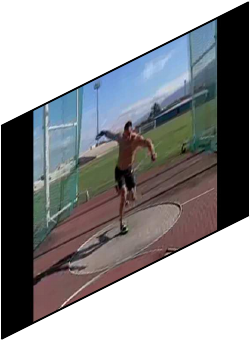} \end{minipage}
 \begin{minipage}{0.058\textwidth} \centering \includegraphics[width=1.00\textwidth]{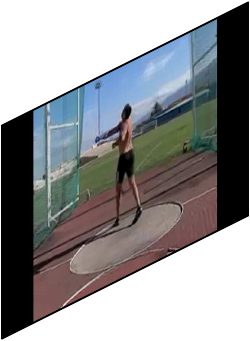} \end{minipage}
 \begin{minipage}{0.058\textwidth} \centering \includegraphics[width=1.00\textwidth]{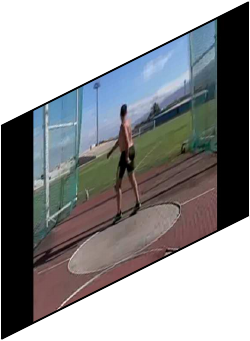} \end{minipage}
 \begin{minipage}{0.058\textwidth} \centering \includegraphics[width=1.00\textwidth]{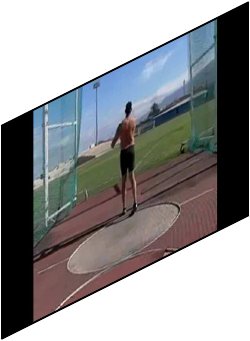} \end{minipage}
 \begin{minipage}{0.058\textwidth} \centering \includegraphics[width=1.00\textwidth]{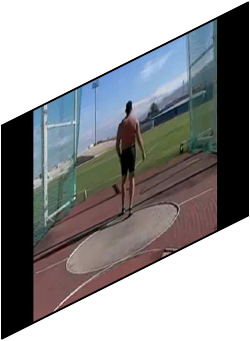} \end{minipage}
 \begin{minipage}{0.058\textwidth} \centering \includegraphics[width=1.00\textwidth]{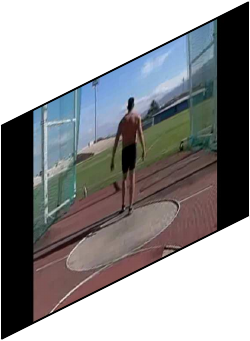} \end{minipage}
 \begin{minipage}{0.058\textwidth} \centering \includegraphics[width=1.00\textwidth]{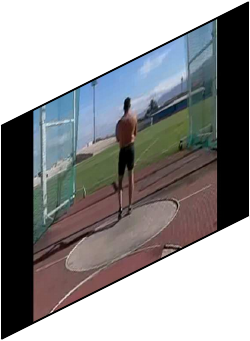} \end{minipage}
 \begin{minipage}{0.058\textwidth} \centering \includegraphics[width=1.00\textwidth]{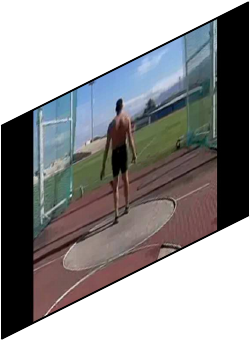} \end{minipage}
 \begin{minipage}{0.058\textwidth} \centering \includegraphics[width=1.00\textwidth]{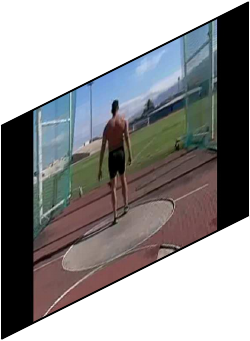} \end{minipage}
 \begin{minipage}{0.058\textwidth} \centering \includegraphics[width=1.00\textwidth]{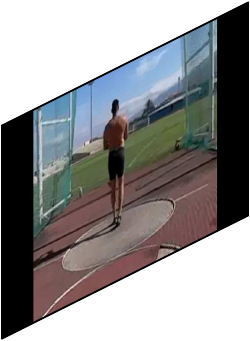} \end{minipage}
 \begin{minipage}{0.058\textwidth} \centering \includegraphics[width=1.00\textwidth]{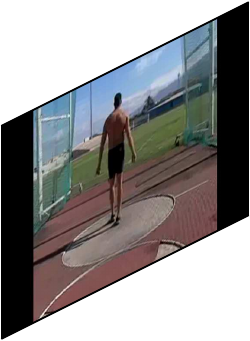} \end{minipage}
 \begin{minipage}{0.058\textwidth} \centering \includegraphics[width=1.00\textwidth]{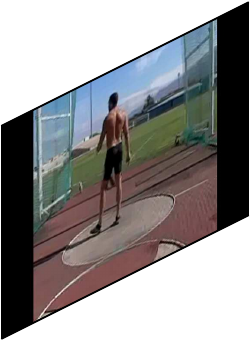} \end{minipage}
 \begin{minipage}{0.058\textwidth} \centering \includegraphics[width=1.00\textwidth]{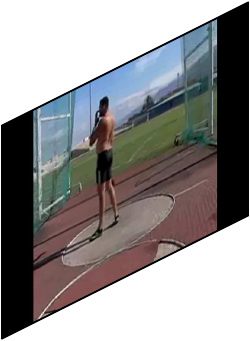} \end{minipage}
 \\ \vspace{2mm}
 \includegraphics[width=1.00\textwidth]{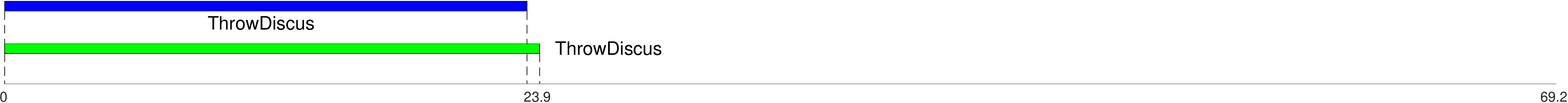}
 \\
 (d)
 \caption{\small Additional qualitative examples of the top localized actions
on THUMOS'14. Each consists of a sequence of frames sampled from a full test
video, the ground-truth (blue) and predicted (green) action segments and class
labels, and a temporal axis showing the time in seconds.}
 \label{fig:additional-2}
\end{figure*}

Besides Fig.~\ref{fig:qualitative}, we show additional qualitative examples on
THUMOS'14 in Fig.~\ref{fig:additional-1} and~\ref{fig:additional-2}. Our
approach successfully localizes the actions in most cases. The failure cases
include: (1) inaccurate boundaries, e.g. Fig.~\ref{fig:qualitative} (middle),
(2) misclassified actions, e.g. Fig.~\ref{fig:qualitative} (bottom) and Fig.~\ref{fig:additional-1} (a), (3) false positives due to indistinguishable
body motions, e.g. Fig.~\ref{fig:additional-1} (b) and Fig.~\ref{fig:additional-2} (c), and (4) false negatives due to small objects and occlusion, e.g. Fig.~\ref{fig:additional-1} (d).

\begin{table}[t]
 \centering
 \begin{tabular}{lc}
  \hline \TBstrut
  \multirow{2}{*}{}                & InceptionV3 RGB        \\
                                   & (ImageNet pre-trained) \\
  \hline \Tstrut
  Zhao et al.~\cite{zhao:iccv2017} &  18.3                  \\ \Bstrut
  Ours                             &  \textbf{26.0}         \\
  \hline
 \end{tabular}
 \vspace{-2mm}
 \caption{\small Action localization mAP (\%) on THUMOS'14 using InceptionV3.
The result of~\cite{zhao:iccv2017} is copied from~\cite{yjxiong:code}.}
 \label{tab:map}
\end{table}

\subsection{Benchmarks using InceptionV3}

Besides I3D features, we also evaluate our method with features extracted from
an InceptionV3 model pre-trained on ImageNet. This provides an apples-to-apples
comparison with the result of Zhao et al.~\cite{zhao:iccv2017} reported
in~\cite{yjxiong:code}. Tab.~\ref{tab:map} shows the action localization mAP on
THUMOS'14. Our approach outperforms Zhao et al.~\cite{zhao:iccv2017} by 7.7\%
in mAP, validating the effectiveness of our proposed architecture.

\subsection{Computational Cost}

Tab.~\ref{tab:time} shows a running time breakdown during the test time for the
following three steps: (1) optical flow extraction, (2) I3D feature extraction,
and (3) proposal and classification. All these running time experiments are
performed on CPUs, so further speedup is possible with GPU devices. The
computational bottleneck is on the optical flow extraction (i.e. 239 ms per
frame).

\begin{table}[t]
 \centering
  \begin{tabular}{l||rl}
  \hline \TBstrut
  Step                      & \multicolumn{2}{c}{Running Time (ms)} \\
  \hline \Tstrut
  Optical Flow              &  239 & per frame          \\ 
  I3D Features              &  825 & per 16 frame input \\ \Bstrut
  Proposal + Classification &  9   & per 3000 frames    \\
  \hline
 \end{tabular}
 \vspace{-2mm}
 \caption{\small Running time (ms) of each step during test time.}
 \label{tab:time}
\end{table}

\end{document}